\documentclass{article}
\usepackage{ijcai26}

\usepackage{times}
\usepackage{soul}
\usepackage{url}
\usepackage[hidelinks]{hyperref}
\usepackage[utf8]{inputenc}
\usepackage[small]{caption}
\usepackage{graphicx}
\usepackage{subfig}
\usepackage{amsmath}
\usepackage{amsthm}
\usepackage{amsfonts}
\usepackage{booktabs}
\usepackage{multirow}
\usepackage[ruled,linesnumbered,vlined]{algorithm2e}
\usepackage{setspace}
\usepackage[switch]{lineno}
\usepackage{color}
\usepackage[normalem]{ulem}

\title{LBA: Textual Hard-Label Adversarial Attack under Low Query Budgets}

\author{
Shixin Guo$^1$ \and
Ming Zhong$^1$ \and
Xuhong Zhang$^2$ \and
Dandan Zhao$^1$ \and
Zhe Wang$^1$ \and
Bo Zhang$^3$ \and
Shouling Ji$^2$ \and
Hao Peng$^1$ \thanks{Corresponding author.}   \\
\affiliations
$^1$Zhejiang Normal University\\
$^2$Zhejiang University\\
$^3$China Electric Power Research Institute\\
\emails
\{spacebound, zhongming, ddzhao, wangzhe97, hpeng\}@zjnu.edu.cn,
\{zhangxuhong, sji\}@zju.edu.cn,
bobozhangiit@gmail.com
}

\begin{document}

\maketitle

\begin{abstract}
Generating high-quality adversarial texts with low query budgets remains a challenging problem in the hard-label scenario. Most existing approaches rely on greedy algorithms, where one position in the text is selected for substitution, followed by the substitutions of other positions. This local search approach may fail to discover high-quality adversarial examples and often leads to excessive query costs. Ideally, an optimal adversarial sample would consider all possible position combinations in the text, but exhaustive search is computationally impractical. To address this challenge, we propose a sampling-based method called LBA, which constructs an approximate distribution of high-quality adversarial examples by integrating both prior and posterior knowledge, and utilizes this distribution for sampling. As sampling progresses, posterior knowledge updates the approximate distribution, which in turn guides more effective sampling.
Extensive experiments on six language models, ranging from small-scale to large-scale architectures across four datasets, demonstrate that LBA significantly outperforms state-of-the-art baselines on all evaluation metrics. Additionally, LLM-based assessment indicates that LBA generates more semantically preserved and comprehensible adversarial texts.
\end{abstract}

\section{Introduction}
Deep neural networks (DNNs) have achieved remarkable success and are widely applied to natural language processing (NLP) tasks\cite{brown2020language,hu-etal-2022-deep,zhangicml23}. However, as DNNs become increasingly popular and widely applied, potential security problems have also raised significant concerns.
It has been observed that DNNs are easily deceived into producing wrong predictions by well-designed malicious examples, known as adversarial attacks\cite{GoodfellowSS14,Sze2014intriguing}.
To reveal the vulnerabilities and improve the robustness of neural language models, researchers have focused on generating plausible adversarial examples.
In recent years, there has been a surge in textual adversarial attacks\cite{alzantot2018generating,linyang2020bert,zang2020word-level}, further intensifying the pursuit of crafting high-quality adversarial texts. 

Based on the outputs the adversary obtained from victim models, existing works on black-box textual adversarial attacks can be categorized into soft-label attacks and hard-label attacks.
Soft-label attacks\cite{li2019textbugger,jin2020bert,zang2020word-level}  typically calculate the importance of words using the confidence scores output by target models, and then modify words based on the importance ranking until the predicted label is reversed. 
Since most real-world APIs do not provide confidence scores, hard-label attacks that only require the top-1 label are more practical and challenging.
The prevalent hard-label attacks usually adopt one of the following algorithms: (1) heuristic algorithms\cite{maheshwary2020}, (2) gradient estimation algorithms\cite{ye2022texthoaxer,ye2022leapattack}, or (3) direction estimation algorithms\cite{liu2024hqa} to optimize an initialized adversarial example.
In real-world applications, query budgets tend to be limited by monetary costs, and a massive number of similar inputs also increases the risk of exposing the adversary.
Therefore, generating high-quality adversarial texts under low query budgets has become a critical concern for hard-label attacks.
Although existing methods strive to improve query efficiency, they still struggle with the trade-off between query efficiency and quality.
To address this issue, we reveal the limitations of existing hard-label attacks and propose a novel approach.

Existing methods essentially follow a greedy search paradigm, leading to query inefficiency. Their exploration strategy iteratively refines adversarial examples, where each iteration improves upon the result of the previous one to improve quality.
To accomplish this, greedy search selects a word position for substitution and locks in the change, then proceeds to the next candidate position without revisiting previous decisions.
However, generating high-quality adversarial examples depends on selecting critical positions and appropriate word substitutions for these positions, which involves exploring substitutions for all combinations of perturbable positions.
During greedy search, the search space of perturbable position combinations gradually shrinks, blocking the discovery of certain feasible solutions. 
Ideally, an optimal adversarial sample would consider all possible position combinations in the text.
However, this would introduce a greater challenge of combinatorial explosion.

To address this challenge, we propose a novel method named LBA, which performs sampling from an approximate distribution of high-quality adversarial examples.
Sampling from a distribution that statistically abstracts the population and theoretically covers all the combinations offers an effective solution for mitigating the explosion of the search space, compared to directly selecting from the population itself.
However, it is not trivial to accurately characterize the distribution of high-quality adversarial examples. To overcome this challenge, we draw inspiration from Metropolis-Hastings (MH) sampling\cite{chib1995understanding} to construct an approximating distribution (i.e., the target function) and sample from it, aiming to obtain samples conforming to the desired distribution.
For effective approximation, we leverage both prior and posterior knowledge to shape the target function. The prior is derived from predefined quality constraints (e.g., perturbation rate and semantic similarity).
During each attack, the posterior is learned from samples accumulated over previous sampling iterations and is used to iteratively update LBA's understanding of aggressiveness, uncovering which positional changes most strongly induce label inversion.
Specifically, we quantify aggressiveness by the importance of each position, measured as its empirical label-flip frequency from adversarial examples discovered in previous iterations.
By leveraging this target function, LBA progressively refines its approximation of the desired distribution through iterative sampling, and is self-correcting even when initial priors are too coarse to accurately characterize the distribution of high-quality adversarial examples.

The sampling process is an iterative cycle of generating a candidate sample and evaluating it for acceptance or rejection. At each iteration, we first generate a candidate by replacing a randomly chosen word in the sample from the previous iteration with a synonym, following the principle of with-replacement sampling.
Subsequently, we incorporate the target function and a transition proposal to probabilistically determine whether to accept and then query the candidate.
As iterations progress, query-derived posterior knowledge refines the estimated importance of positions in the target function, leading to more efficient sampling. In turn, improved sampling yields more informative observations, creating a positive feedback loop.
To sum up, our contributions are as follows:
\begin{itemize}

\item
We reveal the limitations of existing hard-label textual adversarial attacks. Furthermore, to the best of our knowledge, we are the first to formulate query-efficient adversarial text generation as a distribution-based sampling problem.

\item
% 高效的采样策略体现出来
We propose LBA (\textbf{L}ow-query \textbf{B}udget hard-label \textbf{A}ttack), a novel method for generating high-quality adversarial texts in the hard-label scenario.

\item
We conduct experiments on six language models, ranging from small-scale to large-scale architectures, including BERT, LLaMA 2, and GPT-4o, across four datasets. The results demonstrate that LBA significantly outperforms existing hard-label baselines.

% \item
% We utilize GPT-4o to further evaluate the utility of the generated adversarial texts, and the results show that the adversarial texts generated by LBA are more semantically preserved and more comprehensible.
\end{itemize}

\section{Related Work}
\subsection{Soft-Label Adversarial Attack}
% 对抗样本
Soft-label adversarial attacks rely on the confidence scores of target models to craft adversarial examples. Typically, these methods \cite{li2019textbugger,jin2020bert,li2021contextualized,maheshwary2021strong,garg2020bae,gao2024semantic} calculate the importance of words based on confidence scores and then perturb those words according to their importance ranking until adversarial examples are generated.
For instance, Textfooler\cite{jin2020bert} uses importance ranking to sequentially perform synonym substitutions. Additionally, Optimization-based methods\cite{alzantot2018generating,zang2020word-level} utilize combinatorial optimization techniques to craft adversarial texts.
For instance, PSO\cite{zang2020word-level} treats adversarial text generation as a combinatorial optimization problem and adopts a particle swarm optimization algorithm to address the challenge.

\subsection{Hard-Label Adversarial Attack}
% Under the black-box hard-label setting, the decision-based strategy\cite{cheng2019query,chen2020hopskipjumpattack}, which gradually approaches the decision boundary by optimizing an initial example to obtain adversarial examples, is widely used in both computer vision and NLP domains.
Existing hard-label attacks typically adopt heuristic algorithms \cite{maheshwary2020,peng2023textcheater}, gradient estimation-based search \cite{ye2022texthoaxer,ye2022leapattack,LIU2024106461}, or direction estimation-guided search \cite{liu2024hqa} to approach the decision boundary of the target model. HLA\cite{maheshwary2020} explores hard-label adversarial attacks in NLP by employing the genetic algorithm to optimize adversarial texts. However, it is easy to fall into the local optimum and requires numerous queries to maintain its population.
TextHoaxer \cite{ye2022texthoaxer} first formulates the hard-label adversarial attack as a gradient-based optimization problem in the word embedding space. It introduces a perturbation matrix to estimate the gradient at each position, thereby determining the fixed substitute word for each position.
To mitigate inaccurate gradient estimation, LeapAttack \cite{ye2022leapattack} implements an interchange between discrete substitutions and continuous vectors, thereby enhancing the effectiveness of gradient utilization in discrete text.
To alleviate the issue of gradient estimation consuming numerous queries, HQA-Attack \cite{liu2024hqa} evaluates position importance and sequentially predicts the perturbation direction for each position, representing each direction with a transition synonym to avoid querying the entire synonym set. However, the reliance on the greedy search paradigm limits the potential of existing methods to further enhance query efficiency.
In contrast to the aforementioned textual hard-label methods, LBA adopts sampling-based strategy as its search paradigm to efficiently explore the search space.

\begin{figure*}[t]
\centering
\includegraphics[width=0.75\textwidth]{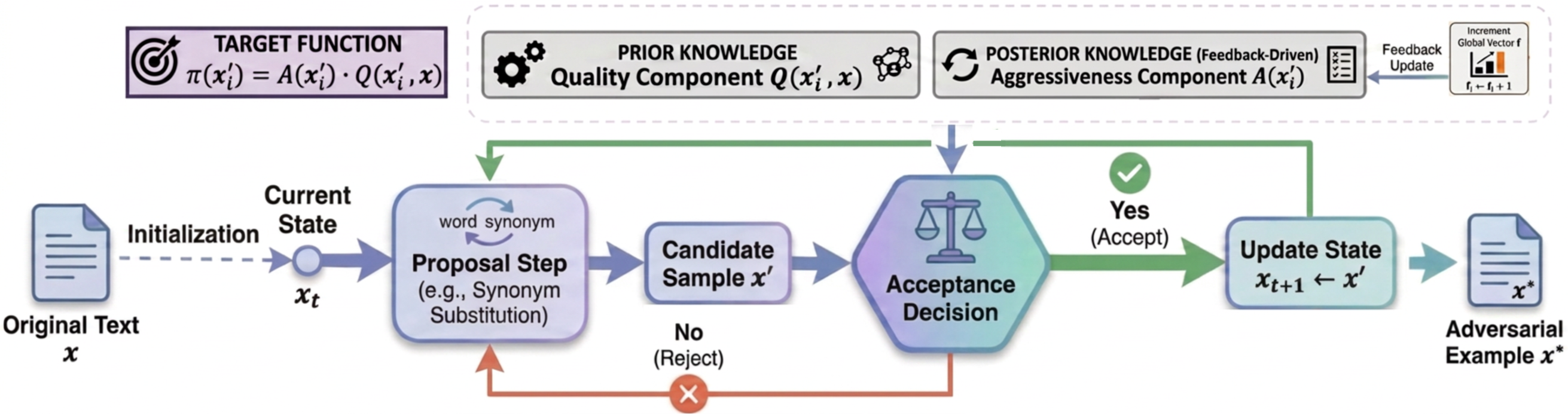}
\caption{Pipeline of the sampling-based method LBA, which leverages both prior and posterior knowledge to shape the target function.}
\label{overview}
\end{figure*}

\section{Proposed Attack}
\subsection{Problem Formulation}
In this paper, we focus on adversarial text generation under the black-box hard-label setting, where the adversary can only access the top-1 predicted label from the target language model.
The target model, denoted by a function $f$, accepts an input text $x = [w_1, w_2, ..., w_n]$ with $n$ words and outputs a predicted label $f(x) = y$, which matches the corresponding ground-truth label. The goal of the attack is to craft a high-quality adversarial text $x^{\prime}$ by synonym substitution in $x$ under a query budget of $\eta$.
Specifically, $x^{\prime}$ is designed to mislead the victim model $f$ into outputting an incorrect prediction result, i.e., $f(x) \neq f(x^{\prime})$.
The term ``high-quality” denotes adversarial texts characterized by high similarity and fluency, which correspond to low perturbation rates, high semantic similarity, and fluent sentences.
These attributes are crucial to aligning with the definition of adversarial texts, which requires them to be imperceptible to humans.
Formally, we define the optimization problem as:
\begin{equation}
\begin{split}
&\arg \max _{x^{\prime}} \operatorname{Q}\left( x^{\prime}, x\right)\\
& s.t. \quad 
\left\{\begin{array}{lc}
Q(x^{\prime}, x)\propto F(x^{\prime}) + S(x^{\prime}, x)\\
f(x)\ne f(x^{\prime}) \\ 
Qrs(x^{\prime}) \le \eta \\
\end{array}\right.
\end{split}
\label{eq1}
\end{equation}
where $Q(x^{\prime}, x)$ denotes the quality of $x^{\prime}$, which is positively correlated with the fluency $F(x^{\prime})$ and similarity $S(x^{\prime}, x)$ of the adversarial example. $Qrs(x^{\prime})$ represents the number of queries required to craft $x^{\prime}$, and $\eta$ is the given query budget.

\subsection{Framework of LBA}
The proposed approach is summarized in Algorithm \ref{algorithm},  which first utilizes random initialization to generate an initial adversarial example $x^{\prime}_0$ and then optimizes it using the sampling-based strategy.

\subsubsection{Initialization}
To establish a starting point for sampling, we use random initialization, an approach commonly used in existing hard-label adversarial attacks\cite{maheshwary2020,ye2022leapattack,ye2022texthoaxer,liu2024hqa}, to create an initial adversarial example.
This process begins by randomly selecting a word $w_j$ from the input text $x$ and substituting it with a randomly chosen synonym from its synonym set $Syn(w_j)$.
This substitution is repeated for other words in $x$ until an example $x^{\prime}_0 = [w^{\prime}_1, w^{\prime}_2, ..., w^{\prime}_n]$ is generated, such that $f(x) \neq f(x^{\prime}_0)$.
Before substituting a word, synonyms with mismatched part-of-speech (POS) tags are filtered out to improve grammatical consistency. To further improve the semantic similarity of the initial adversarial text, we follow previous attacks to substitute some original words back into $x^{\prime}_0$. 
Under limited query budgets, fast random initialization is a practical choice. Moreover, compared to greedy methods that progressively shrink perturbable positions during iteration, ourmethod is not more sensitive to initialization. 
% Specifically, we first determine the substitution order based on the semantic similarity improvement contributed by each original word. Following this order, we iteratively put original words back while ensuring that the example still satisfies the adversarial condition.

\subsubsection{Sampling-Based Strategy}
% We now introduce how to utilize the sampling-based strategy, inspired by MH sampling, to explore the search space, starting from the initial adversarial example $x^{\prime}_0$.
% When the desired distribution is intractable, MH sampling enables obtaining samples that conform to the desired distribution by constructing and sampling from an approximating distribution.
% Notably, MH sampling provides theoretical convergence guarantees under ergodicity, and our adaptive target function naturally satisfies the diminishing adaptation condition as statistics accumulate.
% The sampling is an iterative process consisting of the following steps:
As shown in Figure \ref{overview}, we employ a sampling-based strategy inspired by MH sampling to explore the search space from the initial adversarial example. MH sampling enables drawing samples from an intractable distribution by constructing a tractable approximation. It provides theoretical convergence guarantees under ergodicity, a condition our adaptive target function naturally satisfies as statistics accumulate. The sampling is an iterative process consisting of the following steps:
\begin{itemize}
\item Propose a candidate sample by substituting the word at a randomly selected perturbable position in the current sample (initially set to $x^{\prime}_0$) with a synonym.

\item Calculate the acceptance rate of the candidate sample based on the target function and transition proposal.

\item Accept or reject the candidate: if rejected, retain the current sample; if accepted, query the victim model to validate aggressiveness and update the current sample accordingly.
\end{itemize}
In the following, we provide a detailed introduction to the three steps of sampling:

\begin{algorithm}[t]
\caption{LBA}\label{algorithm}
\label{alg1}
\SetKwInOut{Input}{Input}\SetKwInOut{Output}{Output}
\Input{Original text $x = [w_1, w_2, ..., w_n]$, ground-truth label $y$, target model $f$.}
\Output{Adversarial example $x^\prime$.}
\BlankLine
% \tcp{Initialization}

Generate an initial adversarial example $x^{\prime}_0 = [w^{\prime}_1, w^{\prime}_2, ..., w^{\prime}_n]$ by randomly substitution

Current example $x^{\star} \gets x^{\prime}_0$, optimal adversarial example  $x^\prime \gets x^{\prime}_0$, sampling counter $i \gets 0$

\While {Query budget is not exhausted}
{
% \tcp{Proposing a Candidate Sample}

$i \gets i + 1$, select an index $j$ form perturbable position indices $l_p$

Propose a candidate sample $x^{\prime}_i$ by substituting the $j$-th word of $x^{\star}$ with a synonym $s^k_j$

% \tcp{Calculating the Acceptance Rate}

$\alpha(x^{\prime}_i,x^{\star})$ $\gets$ Calculate the acceptance rate of $x^{\prime}_i$ via Eq. (\ref{eq-ap})

% \tcp{Updating the Adversarial Example}

\If {$\text{Uniform}(0, \beta) < \alpha(x^{\prime}_i,x^{\star})$}{$y_i \gets$ obtain the prediction result of $x^{\prime}_i$

\If {$y_i \ne y$}{
Further improve the quality of $x^{\star}$
}
Updating current example $x^{\star}$ and optimal adversarial example  $x^{\prime}$
}
}
\textbf{Return} $x^{\prime}$ 
\end{algorithm}
\textbf{(1) Proposing a Candidate Sample.}
This step aims to generate a candidate sample from the search space composed of all perturbable positions.
In existing methods, the search space of positions gradually shrinks, which can block the discovery of certain feasible solutions. To avoid this, we generate candidate samples using with-replacement modification over the perturbable positions.
Let $l_p$ denote the set of perturbable positions, i.e., the indices where the words differ between the initial adversarial example and its original counterpart.
We generate the candidate sample $x^{\prime}_i$ by substituting the $j$-th word $w_j$ in the current example $x^{\star}$ with a synonym $s^k_j$ that is randomly picked from $Syn(w_j) = \{s^1_j, s^2_j, ..., s^r_j\}$.
The index $j$ is randomly picked from the set $l_p$. 
There are two reasons for selecting indices from $l_p$.
As shown in previous work, if the perturbable positions are expanded to include all words, the search space becomes excessively large, making it difficult to achieve effective coverage.
Furthermore, these positions, compared to unchanged words, have already been validated for effectiveness during the initialization process.

\textbf{(2) Calculating the Acceptance Rate.}
To select samples that better align with the desired distribution from the proposed candidates and to control the stochasticity of sampling transitions, we use the acceptance rate to determine whether a proposed sample should be accepted.
The acceptance rate is calculated by comparing the relative probabilities of the candidate sample and the current sample with respect to the target function and transition proposal.
Specifically, the target function is designed to approximate the distribution of high-quality adversarial examples.
The accuracy of the distribution approximation directly influences the effectiveness of the obtained samples and plays a guiding role in the overall sampling process.
Additionally, the transition proposal is asymmetric and is measured by the change in semantic similarity between the current sample and the candidate sample, assisting the search in navigating toward semantically coherent adversarial examples.
The acceptance rate of a proposed candidate $x^{\prime}_i$, is defined as:
\begin{equation}
\begin{split}
\alpha (x^{\prime}_i,x^{\star}) = min \left(1, \frac{\pi(x^{\prime}_i)g(x^{\star} \mid x^{\prime}_i)}{\pi(x^{\star})g(x^{\prime}_i\mid x^{\star} )}\right)
\end{split}
\label{eq-ap}
\end{equation}
where $\pi(x^{\prime}_i)$ denotes the target function, and $g(x^{\star} \mid x^{\prime}_i)$ represents the transition proposal. A detailed explanation of these components is provided below.

The target function $\pi(x^{\prime}_i)$ is designed for approximating the distribution of high-quality adversarial examples. Methodologically, a task-driven approach that maximizes the use of knowledge yields a more accurate approximation. Accordingly, the target function incorporates two key attributes, namely aggressiveness and quality, to capture the essential characteristics of high-quality adversarial examples.
Aggressiveness reflects the attack capability of inducing label inversion, informed by posterior knowledge learned from samples accumulated over previous sampling iteration.
Quality measures the similarity between candidate examples and original inputs, as well as fluency, derived from predefined quality constraints.
Specifically, $\pi(x^{\prime}_i)$ is defined as:
\begin{equation}
\begin{split}
\pi(x^{\prime}_i) = A(x^{\prime}_i) \cdot Q(x^{\prime}_i ,x) 
\end{split}
\end{equation}
where $A(x^{\prime}_i)$ and $Q(x^{\prime}_i ,x)$ quantify the aggressiveness and quality, respectively. 

To achieve the estimation of aggressiveness, we transform the statistical information from query feedback during the sampling history into position importance scores.
Previous studies\cite{jin2020bert} have shown that only some key words act as influential signals for the final prediction results.
In the absence of confidence scores, posterior knowledge obtained from previous queries can be utilized to identify key positions more strongly correlated with aggressiveness.
Therefore, we employ the empirical label-flip frequency to measure the importance of word position.
Specifically, the importance of a position is determined by the frequency at which its perturbation induces label inversion, with this frequency calculated based on all discovered adversarial examples.
These frequencies are normalized into final importance scores.
The aggressiveness component is defined as follows:
\begin{equation}
\begin{split}
& A(x^{\prime}_i) = softmax(\mathbf{f})_j = \frac{\exp(f(j))}{\sum_{t=1}^n \exp(f(t))} \\
& \mathbf{f} = [f(1), f(2),..., f(j), f(n)]
\end{split}
\end{equation}
where $f(j)$ represents the number of times that position $j$, the index selected for substitution in the previous step, has appeared among the indices that induce label inversion.

The other component of the target function is the quality $Q(x^{\prime}, x)$. According to the definition in Equation \ref{eq1}, "high-quality" denotes adversarial texts characterized by high similarity and fluency. We evaluate similarity using semantic similarity and perturbation rate, and assess fluency using the increase in grammatical errors and perplexity. Since the priorities among these features differ, we integrate these four metrics via weighted summation, and it is defined as:
\begin{equation}
\begin{split}
Q(x^{\prime}_i, x) = \lambda_1  \mathbb{S}(x^{\prime}_i, x) + \lambda_2 \mathbb{C}(x^{\prime}_i, x^{\prime}_0) + \lambda_3 \mathbb{G}(x^{\prime}_i)+ \lambda_4 \mathbb{P}(x^{\prime}_i, x)
\end{split}
\end{equation}
where $\lambda_1$, $\lambda_2$, $\lambda_3$ and $\lambda_4$ are weight parameters.
$\mathbb{S}(x^{\prime}_i, x)$ measures the semantic cosine similarity between $x^{\prime}_i$ and $x$ as follows:
\begin{equation}
\begin{split}
\mathbb{S}(x^{\prime}_i, x) = cos\left(Enc(x^{\prime}_i), Enc(x)\right)
\end{split}
\end{equation}
where $Enc( \cdot )$ is the universal sentence encoder\cite{cer2018universal}, a machine learning model that converts sentences into high-dimensional vectors.
The term $\mathbb{C}(x^{\prime}_i, x^{\prime}_0)$ quantifies the reduction in perturbation rate, with larger values indicating a smaller perturbation rate in $x^{\prime}_i$ compared to the initial adversarial example $x^{\prime}_0$.
$\mathbb{G}(x^{\prime}_i)$ evaluates grammatical error changes, calculated with the help of LanguageTool, where $\mathbb{G}(x^{\prime}_i)$ is set to 0 if $x^{\prime}_i$ contains more grammatical errors than $x^{\star}$, and 1 otherwise.
The perplexity term $\mathbb{P}(x^{\prime}_i, x)$ calculates the perplexity of $x^{\prime}_i$ and is defined as:
\begin{equation}
\begin{split}
\mathbb{P}(x^{\prime}_i, x) = min \left(2, \frac{ppl(x^{\prime}_i)-ppl(x)}{ppl(x^{\prime}_0)-ppl(x)} \right)
\end{split}
\end{equation}
where $ppl(x)$ is calculated with the help of GPT-2.

The transition proposal $g(x^{\star} \mid x^{\prime}_i)$, representing the transition probability between the current sample and the candidate sample, is defined as:
\begin{equation}
\begin{split}
g(x_i\mid x^{\star}), g(x^{\star}\mid x^{\prime}_i)= softmax \left(1-\mathbb{S}(x^{\star},x),1-\mathbb{S}(x^{\prime}_i,x)\right)
\end{split}
\end{equation}
The reason for choosing semantic similarity as the primary element of the transition proposal is that it plays a crucial role in guiding the search toward the decision boundary while avoiding abrupt jumps. To some extent, previous works\cite{maheshwary2020,liu2024hqa} view the search process as iteratively optimizing the semantic similarity between the original example and the adversarial example.

\textbf{(3) Updating the Adversarial Example.}
This step determines whether to accept a candidate sample for querying the target model and updating the searched adversarial examples. Given the acceptance rate $\alpha(x^{\prime}_i, x^{\star})$ for $x_i$, we sample a value from a uniform distribution over the interval $(0, \beta)$.
If $\alpha(x^{\prime}_i, x^{\star})$ exceeds this sampled value, the candidate sample is accepted.
Compared to using a fixed threshold, the sampled value introduces randomness that allows LBA to occasionally accept lower-probability candidates, enabling proper exploration of the target distribution.
Upon acceptance, the current example $x^{\star}$ is updated to $x^{\prime}_i$ for the next iteration, and the target model is invoked to obtain its corresponding label $y_i$.
If $y_i$ differs from the ground truth label, $x_i$ is identified as an adversarial example found by our sampling-based strategy.
Finally, to further improve the quality of the adversarial example, we substitute the original words back while ensuring that the example still satisfies the adversarial condition.
\section{Experiments}
\subsection{Datasets and Query budgets}
We randomly select 500 samples from the test sets of four commonly used datasets, including two short-text datasets and two long-text datasets: (1) \textbf{MR}\cite{pang2005seeing}, a movie review dataset; (2) \textbf{AG}\cite{Zhang2015CharacterlevelCN}, a 4-class news classification dataset; (3) \textbf{YELP}\cite{Zhang2015CharacterlevelCN}, a binary sentiment classification dataset; and (4) \textbf{IMDB}\cite{maas2011learning}, another binary classification dataset. We allocate three query budgets, as summarized in Table \ref{tab:budgets}, for each dataset based on the average number of queries required by a strong soft-label attack, Textfooler\cite{jin2020bert}.
In all experiments, all queries, including queries used to initialize adversarial examples in hard-label attacks, are counted in the given budgets.

\subsection{Victim Models}
Based on the classification in \cite{zhao2023survey}, we select six representative language models from three categories as victim models: Neural Language Models (NLMs), Pre-trained Language Models (PLMs), and Large Language Models (LLMs).
For NLMs and PLMs, we follow the setup used in existing hard-label adversarial attacks and target three models: \textbf{WordCNN}\cite{kim2014convolutional}, \textbf{WordLSTM}\cite{s1997long}, and \textbf{BERT}\cite{devlin2019bert}. 
All the NLMs and PLMs are provided by TextAttack \cite{morris2020textattack}.
For LLMs, we adopt the following three popular models as victims: \textbf{LLaMA2-7b-chat} (LLaMA2, \cite{touvron2023llama}), \textbf{Vicuna-7b-v1.5} (Vicuna, \cite{vicuna2023}), and \textbf{GPT-4o-mini-tts-2025-03-20} (GPT-4o, \cite{openai2023gpt}).
For the generation parameters of the two open-source LLMs, i.e., LLaMA2 and Vicuna, we set the temperature to 0.1 and the maximum tokens to 15.

\begin{table}[t]
\caption{Average lengths of datasets and corresponding query budgets.}
\centering
\label{tab:budgets}
\resizebox{0.4\textwidth}{!}{
\begin{tabular}{cccccc}
\hline
\multirow{2}{*}{\textbf{Dataset}} &
  \multirow{2}{*}{\textbf{Avg Len}} &
  \multirow{2}{*}{\textbf{\begin{tabular}[c]{@{}c@{}}Soft-Label\\ Avg Qrs\end{tabular}}} &
  \multicolumn{3}{c}{\textbf{Budgets}} \\ \cline{4-6} 
     &     &     & \textbf{Tiny} & \textbf{Tight} & \textbf{Moderate} \\ \hline
MR   & 20  & 110 & 50            & 100            & 150               \\
AG   & 43  & 255 & 100           & 200            & 300               \\
YELP & 152 & 406 & 200           & 400            & 600               \\
IMDB & 215 & 733 & 400           & 700            & 1000              \\
\hline 
\end{tabular}
}
\end{table}

\begin{table*}[t]
\centering
\caption{Comparison of perturbation rate (Pert.), semantic similarity (Sim.), increase in grammatical errors(Ige.) and perplexity (PPL) with given query budgets when attacking BERT. For a fair comparison, methods with low attack success rates (ASR.) - more than 20\% below other baselines (denoted by underscored values) - are excluded from the quality comparison.}
\label{tab:attack-on-bert}
\resizebox{0.9\textwidth}{!}{
\begin{tabular}{cc|ccccc|ccccc|ccccc}
\hline
\multirow{2}{*}{\textbf{Dataset}}          & \multirow{2}{*}{\textbf{Method}} & \multicolumn{5}{c|}{\textbf{Tiny Budget}}                                                           & \multicolumn{5}{c|}{\textbf{Tight Budget}}                                                          & \multicolumn{5}{c}{\textbf{Moderate Budget}}                                                        \\ \cline{3-17} 
                                           &                                  & \textbf{ASR.(\%)}      & \textbf{Pert.(\%)} & \textbf{Sim.}   & \textbf{Ige.(\%)} & \textbf{PPL}    & \textbf{ASR.(\%)}      & \textbf{Pert.(\%)} & \textbf{Sim.}   & \textbf{Ige.(\%)} & \textbf{PPL}    & \textbf{ASR.(\%)}      & \textbf{Pert.(\%)} & \textbf{Sim.}   & \textbf{Ige.(\%)} & \textbf{PPL}    \\ \hline
\multicolumn{1}{c|}{\multirow{7}{*}{MR}}   & \textit{Textbugger}              & \underline{32.57}        & 10.16              & 0.9268          & 6.34              & 207.03          & \underline{58.92}        & 12.10              & 0.9105          & 6.37              & 212.83          & \underline{63.28}        & 13.32              & 0.9033          & 6.98              & 223.52          \\
\multicolumn{1}{c|}{}                      & \textit{Textfooler}              & \underline{13.69}        & 11.39              & 0.8846          & 0.82              & 211.34          & \underline{51.87}        & 13.90              & 0.8609          & 0.88              & 193.60          & \underline{73.24}        & 14.89              & 0.8514          & 1.08              & 195.86          \\ \cline{3-17} 
\multicolumn{1}{c|}{}                      & HLA                              & \multirow{5}{*}{90.66} & 30.95              & 0.7201          & 2.98              & 427.86          & \multirow{5}{*}{93.57} & 27.17              & 0.7468          & 2.23              & 371.42          & \multirow{5}{*}{94.61} & 24.14              & 0.7751          & 1.88              & 331.31          \\
\multicolumn{1}{c|}{}                      & Texthoaxer                       &                        & 21.56              & 0.7941          & 2.24              & 304.17          &                        & 16.91              & 0.8415          & 1.62              & 243.44          &                        & 15.76              & 0.8525          & 1.46              & 224.70          \\
\multicolumn{1}{c|}{}                      & LeapAttack                       &                        & 23.63              & 0.7737          & 2.30              & 325.37          &                        & 16.96              & 0.8438          & 1.71              & 234.07          &                        & 14.66              & 0.8656          & 1.35              & 217.16          \\
\multicolumn{1}{c|}{}                      & HQA-Attack                       &                        & 25.15              & 0.7614          & 2.31              & 343.57          &                        & 16.23              & 0.8537          & 1.52              & 229.11          &                        & 14.42              & 0.8735          & 1.51              & 208.17          \\
\multicolumn{1}{c|}{}                      & \textbf{LBA}                     &                        & \textbf{14.03}     & \textbf{0.8477} & \textbf{1.47}     & \textbf{209.33} &                        & \textbf{12.41}     & \textbf{0.8691} & \textbf{1.22}     & \textbf{207.07} &                        & \textbf{12.05}     & \textbf{0.8741} & \textbf{0.88}     & \textbf{194.47} \\ \hline
\multicolumn{1}{c|}{\multirow{7}{*}{AG}}   & \textit{Textbugger}              & \underline{23.94}        & 7.64               & 0.9490          & 2.58              & 212.96          & \underline{45.55}        & 13.33              & 0.9169          & 4.19              & 256.12          & \underline{56.78}        & 16.84              & 0.9006          & 5.33              & 291.55          \\
\multicolumn{1}{c|}{}                      & \textit{Textfooler}              & \underline{16.10}        & 6.27               & 0.9567          & 0.16              & 183.03          & \underline{36.23}        & 10.21              & 0.9269          & 0.74              & 223.14          & \underline{55.93}        & 14.38              & 0.9007          & 1.15              & 261.18          \\ \cline{3-17} 
\multicolumn{1}{c|}{}                      & HLA                              & \multirow{5}{*}{77.97} & 39.32              & 0.6691          & 2.61              & 678.43          & \multirow{5}{*}{86.02} & 35.12              & 0.7002          & 2.40              & 594.65          & \multirow{5}{*}{87.71} & 28.98              & 0.7466          & 2.23              & 485.43          \\
\multicolumn{1}{c|}{}                      & Texthoaxer                       &                        & 25.42              & 0.7878          & 1.80              & 475.20          &                        & 18.06              & 0.8398          & 1.07              & 342.72          &                        & 17.71              & 0.8417          & 0.85              & 337.16          \\
\multicolumn{1}{c|}{}                      & LeapAttack                       &                        & 31.76              & 0.7288          & 2.32              & 567.38          &                        & 18.91              & 0.8346          & 1.27              & 346.83          &                        & 15.84              & 0.8617          & 0.96              & 312.46          \\
\multicolumn{1}{c|}{}                      & HQA-Attack                       &                        & 33.65              & 0.7176          & 2.45              & 584.73          &                        & 19.63              & 0.8332          & 1.28              & 366.63          &                        & 14.87              & 0.8648          & 1.03              & 292.64          \\
\multicolumn{1}{c|}{}                      & \textbf{LBA}                     &                        & \textbf{14.30}     & \textbf{0.8501} & \textbf{0.74}     & \textbf{292.53} &                        & \textbf{13.66}     & \textbf{0.8521} & \textbf{0.48}     & \textbf{285.75} &                        & \textbf{12.74}     & \textbf{0.8653} & \textbf{0.35}     & \textbf{278.27} \\ \hline
\multicolumn{1}{c|}{\multirow{7}{*}{YELP}} & \textit{Textbugger}              & \underline{38.40}        & 8.82               & 0.9388          & 3.96              & 120.02          & \underline{63.24}        & 9.23               & 0.9369          & 3.83              & 110.65          & \underline{76.39}        & 9.43               & 0.9374          & 3.89              & 109.06          \\
\multicolumn{1}{c|}{}                      & \textit{Textfooler}              & \underline{32.24}        & 7.67               & 0.9352          & 0.95              & 101.61          & \underline{63.04}        & 8.04               & 0.9374          & 0.89              & 96.48           & 80.08                  & 8.31               & 0.9376          & 0.87              & 97.70           \\ \cline{3-17} 
\multicolumn{1}{c|}{}                      & HLA                              & \multirow{5}{*}{97.13} & 34.22              & 0.6920          & 2.78              & 438.88          & \multirow{5}{*}{99.38} & 25.34              & 0.7772          & 2.59              & 328.51          & \multirow{5}{*}{99.59} & 20.16              & 0.8021          & 2.36              & 210.70          \\
\multicolumn{1}{c|}{}                      & Texthoaxer                       &                        & 17.22              & 0.8582          & 2.25              & 198.83          &                        & 13.68              & 0.8937          & 1.64              & 173.69          &                        & 11.33              & 0.9099          & 1.40              & 139.97          \\
\multicolumn{1}{c|}{}                      & LeapAttack                       &                        & 23.03              & 0.7991          & 2.89              & 281.54          &                        & 17.03              & 0.8619          & 2.22              & 221.51          &                        & 12.55              & 0.9023          & 1.46              & 166.95          \\
\multicolumn{1}{c|}{}                      & HQA-Attack                       &                        & 23.68              & 0.7941          & 2.76              & 283.89          &                        & 16.66              & 0.8695          & 2.02              & 216.36          &                        & 12.93              & 0.9003          & 1.60              & 167.51          \\
\multicolumn{1}{c|}{}                      & \textbf{LBA}                     &                        & \textbf{8.15}      & \textbf{0.9256} & \textbf{1.05}     & \textbf{105.15} &                        & \textbf{7.26}      & \textbf{0.9352} & \textbf{0.79}     & \textbf{103.03} &                        & \textbf{6.59}      & \textbf{0.9424} & \textbf{0.76}     & \textbf{97.46}  \\ \hline
\multicolumn{1}{c|}{\multirow{7}{*}{IMDB}} & \textit{Textbugger}              & \underline{44.78}        & 4.70               & 0.9722          & 2.24              & 99.92           & \underline{69.78}        & 6.54               & 0.9606          & 2.79              & 102.24          & 76.12                  & 7.71               & 0.9521          & 3.22              & 107.08          \\
\multicolumn{1}{c|}{}                      & \textit{Textfooler}              & \underline{40.30}        & 3.06               & 0.9833          & 0.34              & 80.68           & \underline{66.04}        & 4.45               & 0.9733          & 0.49              & 84.43           & 75.00                  & 5.38               & 0.9676          & 0.64              & 88.27           \\ \cline{3-17} 
\multicolumn{1}{c|}{}                      & HLA                              & \multirow{5}{*}{87.31} & 22.87              & 0.8249          & 2.42              & 268.39          & \multirow{5}{*}{88.81} & 19.87              & 0.8549          & 2.12              & 223.39          & \multirow{5}{*}{93.28} & 18.78              & 0.8767          & 2.29              & 198.19          \\
\multicolumn{1}{c|}{}                      & Texthoaxer                       &                        & 14.14              & 0.9028          & 1.67              & 175.68          &                        & 9.99               & 0.9373          & 1.16              & 138.13          &                        & 8.77               & 0.9453          & 1.14              & 133.21          \\
\multicolumn{1}{c|}{}                      & LeapAttack                       &                        & 19.88              & 0.8436          & 2.30              & 230.82          &                        & 10.91              & 0.9233          & 1.22              & 157.60          &                        & 9.28               & 0.9349          & 1.08              & 140.48          \\
\multicolumn{1}{c|}{}                      & HQA-Attack                       &                        & 17.85              & 0.8643          & 2.07              & 211.92          &                        & 13.98              & 0.8985          & 1.72              & 190.54          &                        & 10.14              & 0.9288          & 1.15              & 152.95          \\
\multicolumn{1}{c|}{}                      & \textbf{LBA}                     &                        & \textbf{4.21}      & \textbf{0.9728} & \textbf{0.49}     & \textbf{89.95}  &                        & \textbf{3.89}      & \textbf{0.9752} & \textbf{0.46}     & \textbf{88.93}  &                        & \textbf{3.68}      & \textbf{0.9767} & \textbf{0.47}     & \textbf{87.50}  \\ \hline
\end{tabular}
}
\end{table*}
\begin{figure*}[]
    \centering
    \subfloat[\scriptsize MR (Pert.)\label{fig:imagemb1}]{
        \includegraphics[width=0.21\textwidth, height=2.3cm]{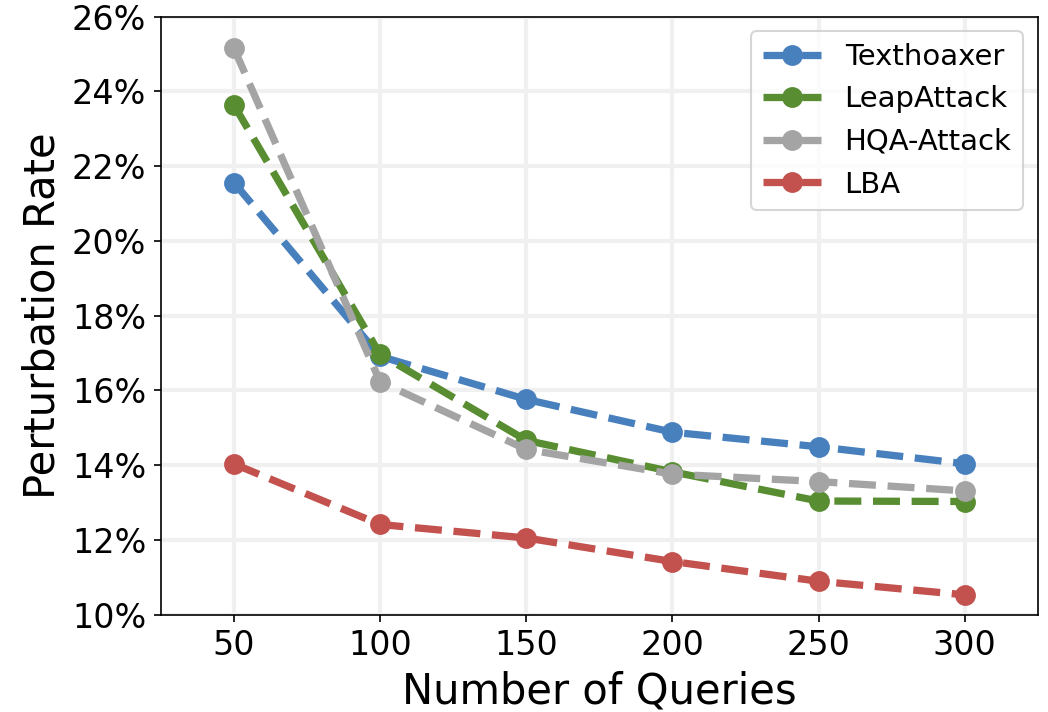}
    }
    \hspace{0.01\textwidth} % 为子图之间添加水平间距
    \subfloat[\scriptsize YELP (Pert.)\label{fig:imagemb2}]{
        \includegraphics[width=0.21\textwidth, height=2.3cm]{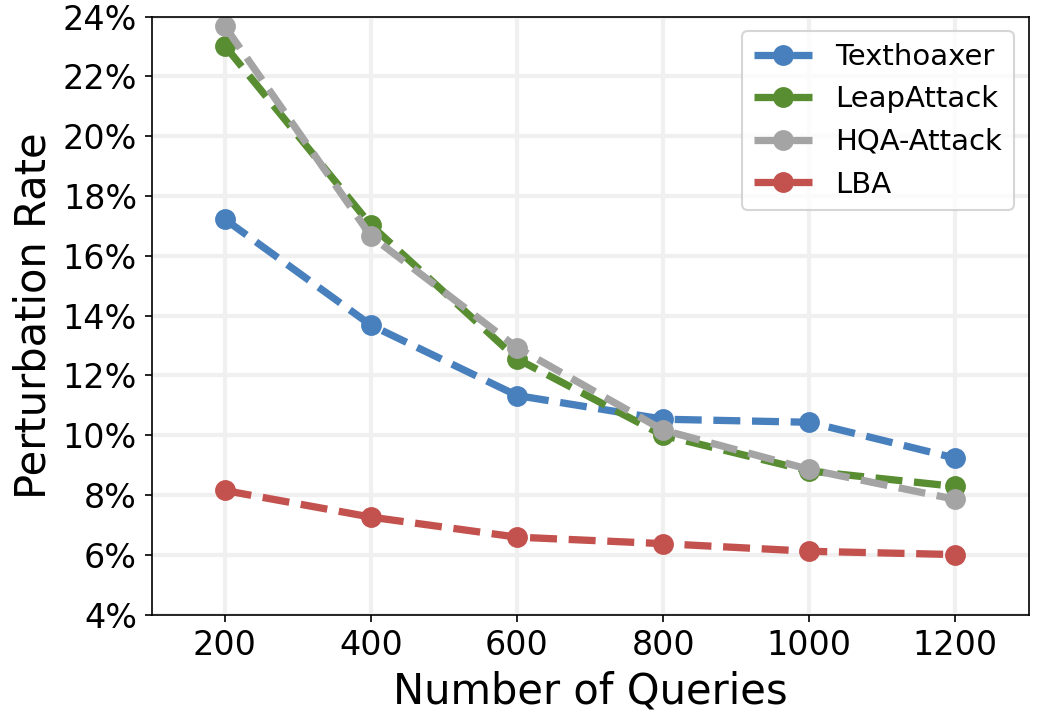}
    }
    \hspace{0.01\textwidth}
    \subfloat[\scriptsize MR (Sim.)\label{fig:imagemb3}]{
        \includegraphics[width=0.21\textwidth, height=2.3cm]{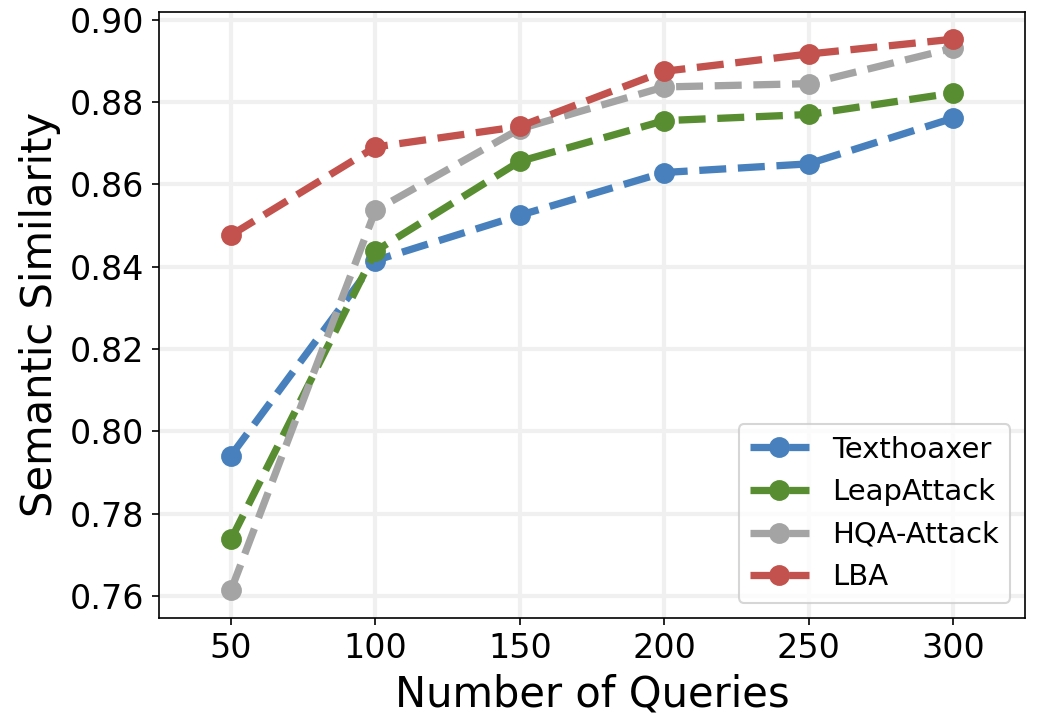}
    }
    \hspace{0.01\textwidth}
    \subfloat[\scriptsize YELP (Sim.)\label{fig:imagemb4}]{
        \includegraphics[width=0.21\textwidth, height=2.3cm]{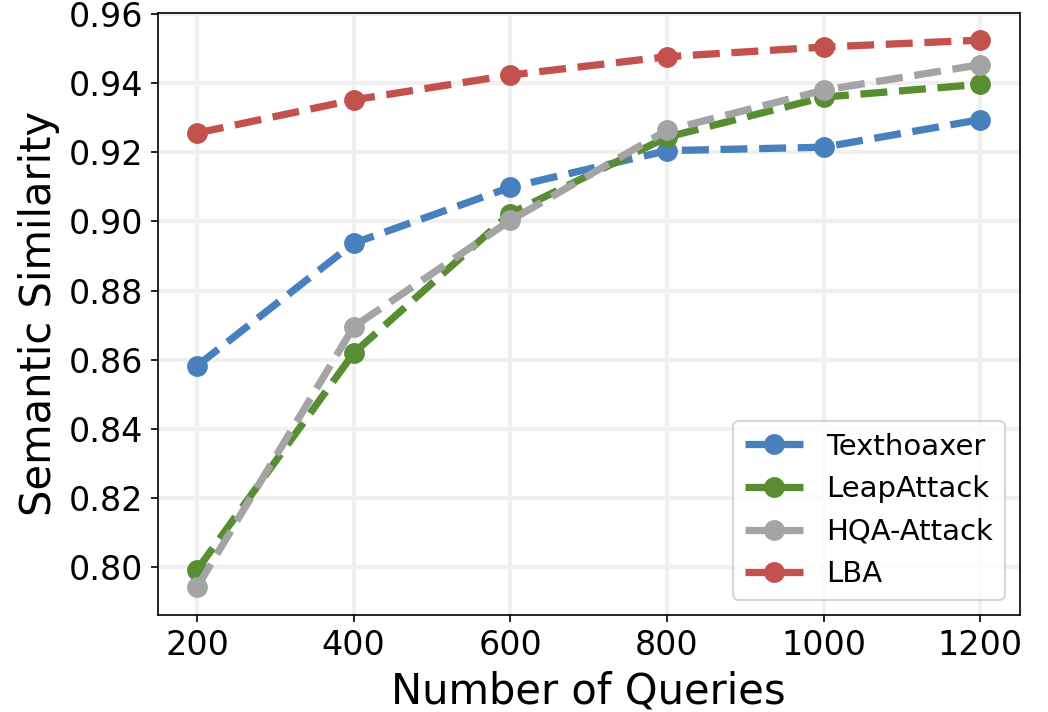}
    }
\caption{Comparison on Pert. (a,b) and Sim. (c,d) in various query budgets for attacks against BERT.}
\label{more-budgets}
\end{figure*}
\subsection{Baselines}
We choose two strong soft-label attacks and four state-of-the-art hard-label attacks as our baselines. These baselines include: (1) \textbf{Textbugger}\cite{li2019textbugger}, a soft-label attack that leverags word importance ranking to guide perturbations; (2) \textbf{Textfooler}\cite{jin2020bert}, a soft-label attack baseline; (3) \textbf{HLA}\cite{maheshwary2020}, a heuristic-based hard-label attack method; (4) \textbf{Texthoaxer}\cite{ye2022texthoaxer}, a gradient-based hard-label attack method; (5) \textbf{LeapAttack}\cite{ye2022leapattack}, another gradient-based hard-label attack method; and (6) \textbf{HQA-Attack}\cite{liu2024hqa}, a semantic similarity-guided hard-label attack method.
We grant the soft-label attacks access to the confidence scores output by the target models to successfully execute the attacks.

\subsection{Implementation Setting}
Following prior work\cite{ye2022leapattack,ye2022texthoaxer,liu2024hqa}, all the hard-label attacks utilize the same initialization method originally proposed by HLA \cite{maheshwary2020} to ensure a fair comparison.
For the quality component weights in our target function, we set $\lambda_1=0.25$, $\lambda_2=0.50$, $\lambda_3=0.15$, and $\lambda_4=0.10$. These values were determined through a sensitivity analysis, where we first optimized the allocation between the similarity group ($\lambda_1+\lambda_2$) and the fluency group ($\lambda_3+\lambda_4$), then tuned the ratio within each group. Detailed parameter tuning results are provided in Appendix C.
% For the weight parameters of the target function in LBA, we set them as follows:$\lambda_1 =0.25 $, $\lambda_2 = 0.5$, $\lambda_3=0.15 $ and $\lambda_4 =0.15$ .
% The weights are set to prioritize critical positions ($\lambda_2 = 0.5$) while balancing other factors. This configuration allows LBA to explore combinations of critical positions more frequently, thereby improving sampling efficiency to some extent.

\subsection{Evaluation Metrics}
We evaluate the quality of adversarial examples from two dimensions: similarity and fluency.
Following previous studies\cite{ye2022texthoaxer,ye2022leapattack,liu2024hqa}, we use two metrics \textbf{perturbation rate} (Pert.) and \textbf{semantic similarity} (Sim.), to quantify the similarity. The semantic similarity is measured by the cosine similarity between the original text and its adversary, calculated using the Universal Sentence Encoder\cite{cer2018universal}.
To assess the fluency of adversarial texts, we employ two metrics: \textbf{perplexity} (PPL) and the \textbf{increase in grammatical errors} (Ige.).
We utilize GPT-2\cite{radford2019language} to calculate the perplexity of texts. The increase in grammatical errors refers to the percentage of increased grammar mistakes compared to the original text.

To more objectively evaluate the aggressiveness and quality of adversarial texts, we further utilize GPT-4o as an evaluator and propose two metrics: \textbf{consistency rate} (CR) and \textbf{fully understood rate} (FUR). The CR metric measures the percentage of adversarial examples classified by GPT-4o as matching their ground truth labels, distinguishing whether aggressiveness stems from altering the original semantics:
\begin{equation}
\begin{split}
CR = \frac{\sum_{x^{\prime}\in X_{adv}}^{} \mathbf{I}(E_{val}(x^{\prime})=y)}{\mid X_{adv}\mid}
\end{split}
\end{equation}
where $X_{adv}$ is the set of adversarial texts, $y$ is the ground truth label of the original text, $E_{val}(x^{\prime})$ denotes GPT-4o's predicted label, and $\mathbf{I}(\cdot)$ returns 1 if the condition holds.
The FUR metric is used to assess whether GPT-4o can fully understand the content of adversarial text, highlighting the robustness of the attack.  A "fully understood" response indicates that the text is coherent and fluent, suggesting that perturbations introduced by the attack are more difficult to detect.
FUR is defined as follows:
\begin{equation}
FUR = \frac{\sum_{x^{\prime}\in X_{adv}} \mathbf{I}(E_{val}(x^{\prime}) = 1)}{|X_{adv}|}
\end{equation}
where $E_{val}(x^{\prime})$ denotes whether GPT-4o's response is classified as "fully understood" (represented by 1).

\subsection{Experimental Results}
\subsubsection{Comparison of Similarity and Fluency}
We performed experiments with the given query budgets against WordCNN, WordLSTM, and BERT. 
Due to limited space, we only present the attack performance against BERT in Table \ref{tab:attack-on-bert}, and present results on other victim models in Appendix \ref{a1}.
As shown in Table \ref{tab:attack-on-bert}, LBA significantly outperforms the baseline methods under tiny, tight, and moderate query budgets.
It achieves the highest similarity (represented by the metrics Pert. and Sim.) and fluency (represented by the metrics Ige. and PPL.) against all victim models across four datasets.
For example, when attacking BERT under the tiny budget, our method reduces the average Pert., Ige., and PPL by 56.18\%, 58.56\%, and 47.76\%, respectively, while improving the average Sim. by 11.74\%, compared to the second-best baseline.
Moreover, it can be observed that on datasets with larger combinatorial spaces, namely the long-text datasets YELP and IMDB, the performance gap in the quality metrics between LBA and other baselines is more significant compared to the other two datasets.
This trend further highlights the effectiveness of LBA's distribution-based sampling strategy in efficiently considering all possible position combinations. To further analyze the distribution of generated samples in terms of similarity, we also compare the cumulative distributions of perturbation rate and semantic similarity for adversarial examples in Appendix \ref{a2}.
As the query budget increases from tiny to tight and moderate levels, the results show that LBA maintains its superiority over other hard-label baselines.
\begin{table*}[t]
\centering
\caption{Comparison of Pert., Sim., Ige. and
PPL with tiny query budgets in attacks against target LLMs.}
\label{tab:attack-on-llms}
\resizebox{\textwidth}{!}{
\begin{tabular}{cc|ccccc|ccccc|ccccc}
\hline
\multirow{2}{*}{\textbf{Dataset}} & \multirow{2}{*}{\textbf{Method}} & \multicolumn{5}{c|}{LLaMA2}                                                                  & \multicolumn{5}{c|}{Vicuna}                                                                 & \multicolumn{5}{c}{GPT-4o}                                                                           \\ \cline{3-17} 
                                  &                                  & \textbf{ASR.(\%)}      & \textbf{Pert.(\%)} & \textbf{Sim.}   & \textbf{Ige.(\%)} & \textbf{PPL}    & \textbf{ASR.(\%)}      & \textbf{Pert.(\%)} & \textbf{Sim.}   & \textbf{Ige.(\%)} & \textbf{PPL}    & \textbf{ASR.(\%)}      & \textbf{Pert.(\%)} & \textbf{Sim.}   & \textbf{Ige.(\%)} & \textbf{PPL}    \\ \hline
\multirow{4}{*}{MR}               & Texthoaxer                       & \multirow{4}{*}{89.58} & 21.03              & 0.7905          & 2.08              & 308.29          & \multirow{4}{*}{96.61} & 14.77              & 0.8502          & 1.21              & 228.04          & \multirow{4}{*}{71.68} & 32.60              & 0.6619          & 2.71              & 440.93          \\
                                  & LeapAttack                       &                        & 25.22              & 0.7461          & 2.54              & 351.06          &                        & 16.34              & 0.8388          & 2.17              & 242.68          &                        & 33.17              & 0.6578          & 2.83              & 442.51          \\
                                  & HQA-Attack                       &                        & 25.06              & 0.7569          & 2.42              & 363.66          &                        & 16.14              & 0.8537          & 1.16              & 231.93          &                        & 31.77              & 0.6809          & 2.69              & 436.88          \\
                                  & \textbf{LBA}                     &                        & \textbf{14.88}     & \textbf{0.8322} & \textbf{1.41}     & \textbf{243.41} &                        & \textbf{11.37}     & \textbf{0.8729} & \textbf{1.09}     & \textbf{206.63} &                        & \textbf{19.78}     & \textbf{0.7855} & \textbf{2.02}     & \textbf{308.33} \\ \hline
\multirow{4}{*}{YELP}             & Texthoaxer                       & \multirow{4}{*}{63.11} & 27.86              & 0.7471          & 3.42              & 349.59          & \multirow{4}{*}{74.69} & 22.96              & 0.8001          & 2.78              & 318.34          & \multirow{4}{*}{50.74} & 33.11              & 0.6939          & 3.48              & 456.67          \\
                                  & LeapAttack                       &                        & 29.25              & 0.7343          & 3.65              & 403.38          &                        & 23.24              & 0.7924          & 2.61              & 317.54          &                        & 34.63              & 0.6859          & 3.75              & 464.63          \\
                                  & HQA-Attack                       &                        & 28.13              & 0.7459          & 3.54              & 346.41          &                        & 23.82              & 0.7930          & 2.77              & 325.70          &                        & 31.35              & 0.7134          & 3.04              & 351.41          \\
                                  & \textbf{LBA}                     &                        & \textbf{12.94}     & \textbf{0.8792} & \textbf{1.54}     & \textbf{123.22} &                        & \textbf{9.76}      & \textbf{0.9096} & \textbf{1.10}     & \textbf{96.80}  &                        & \textbf{22.31}     & \textbf{0.7728} & \textbf{2.58}     & \textbf{238.21} \\ \hline
\end{tabular}
}
\end{table*}
\subsubsection{Comparison of Attack under More Query Budgets}
In real-world scenarios, both limited and ample query budgets may be encountered.
Therefore, to more comprehensively evaluate the applicability of LBA, we further compare the Pert. and Sim. of hard-label attacks across various query budgets.
Considering the effect of text length on the search space, we choose two datasets with varying lengths: the short-text MR dataset and the long-text YELP dataset.
The results shown in Figure \ref{more-budgets} indicate that LBA's performance continues to improve as the query budget increases and always surpasses that of other methods.
This suggests that LBA effectively adapts to both limited and ample query budgets, making it well-suited for real-world applications.
\subsubsection{Comparison of Attacking LLMs}
To further evaluate the practicality of LBA on LLMs, we conduct experiments on representative LLMs under the tiny query budget.
In view of the observed performance on NLMs and PLMs, we exclude HLA and select TextHoaxer, LeapAttack, and HQA-Attack as the baselines. We use 500 texts from two representative datasets, MR (a short-text dataset) and YELP (a long-text dataset), as the target samples.
As shown in Table \ref{tab:attack-on-llms}, the results show that LBA generates the highest quality adversarial texts in all three LLMs.
For example, in attacks on the long-text dataset YELP,
compared with the second best results, LBA achieves a 12.39\% lower perturbation rate, 1.28\% lower increase in grammatical errors, and 185.67 lower perplexity, and an 11.74\% higher semantic similarity compared to the second-best method.
Overall, the results not only confirm the robust practicality of LBA but also highlight the need to enhance the adversarial robustness of LLMs.
\begin{table}[]
\caption{Results of utility evaluation using GPT-4o. Acc stands for GPT-4o prediction accuracy of original texts.} 
\label{tab:gpt4-eval}
\centering
\resizebox{0.35\textwidth}{!}{
\begin{tabular}{c|c|c|c|c}
\hline 
\textbf{Dataset} &
  \textbf{Method} &
  \textbf{Acc(\%)} &
  \textbf{\begin{tabular}[c]{@{}c@{}}CR (\%)\end{tabular}} &
  \textbf{\begin{tabular}[c]{@{}c@{}}FUR (\%)\end{tabular}} \\ \hline
\multirow{5}{*}{MR}   & HLA          & \multirow{5}{*}{91.5} & 60.61          & 28.41          \\
                           & Texthoaxer   &                       & 66.11          & 34.45          \\
                           & LeapAttack   &                       & 65.45          & 33.41          \\
                           & HQA-Attack   &                       & 62.38          & 32.79          \\
                           & \textbf{LBA} &                       & \textbf{66.20} & \textbf{38.03} \\ \hline
\multirow{5}{*}{AG}   & HLA          & \multirow{5}{*}{89.4} & 79.03          & 9.68           \\
                           & Texthoaxer   &                       & 82.06          & 13.74          \\
                           & LeapAttack   &                       & 75.54          & 12.50          \\
                           & HQA-Attack   &                       & 73.16         & 13.13          \\
                           & \textbf{LBA} &                       & \textbf{87.43} & \textbf{15.51} \\ \hline
\multirow{5}{*}{YELP} & HLA          & \multirow{5}{*}{98.4} & 92.62          & 15.57          \\
                           & Texthoaxer   &                       & 93.02          & 25.97          \\
                           & LeapAttack   &                       & 91.97          & 22.41          \\
                           & HQA-Attack   &                       & 92.59          & 24.36          \\
                           & \textbf{LBA} &                       & \textbf{94.67} & \textbf{32.41} \\ \hline
\multirow{5}{*}{IMDB} & HLA          & \multirow{5}{*}{93.3} & 61.02          & 32.20          \\
                           & Texthoaxer   &                       & 88.78          & 46.93          \\
                           & LeapAttack   &                       & 88.89          & 32.20          \\
                           & HQA-Attack   &                       & 89.27          & 49.77          \\
                           & \textbf{LBA} &                       & \textbf{91.88} & \textbf{67.95} \\ \hline
\end{tabular}
}
\end{table}

\subsection{Evaluation by LLM Assistant}
To more accurately and objectively evaluate the utility of the generated adversarial text, we leverage GPT-4o's robust natural language understanding capabilities to automatically assess the adversarial examples.
This evaluation focuses on two key aspects: aggressiveness and quality.
Aggressiveness is measured by CR, which helps distinguish whether the aggressiveness of adversarial examples stems from altering the original meaning, further evaluating the semantic preservation of the samples.
Quality is measured by FUR, which reflects whether the fluency of the samples aligns with natural language, further evaluating the comprehensibility of the samples.
We use the adversarial texts generated against BERT with the tiny query budget.
As shown in Table \ref{tab:gpt4-eval}, LBA achieves the highest CR and FUR among the hard-label baselines, indicating that these adversarial texts remain the most semantically preserved and comprehensible.

\subsection{Human Evaluation}
Although recent studies have shown that GPT-4o-based evaluation correlates well with human judgment, we further conduct a human evaluation to directly validate the utility of generated adversarial examples from a human perspective. Specifically, we evaluate the BERT model using HLA, TextHoaxer, LeapAttack, HQA-Attack, and our LBA on the MR and IMDB datasets under the given tight budget. For each dataset, we randomly select 50 original samples and use each attack method to generate the corresponding 50 adversarial examples. We then invite 10 volunteers to annotate the sentiment labels for these samples and calculate the average classification accuracy. As shown in Table \ref{human_eval}, LBA achieves the highest accuracy on both datasets, further confirming that our method successfully fools the victim model while preserving the original semantics.
\begin{table}[]
\caption{Average classification accuracy(\%) given by human evaluators.}
\centering
\label{human_eval}
\resizebox{0.42\textwidth}{!}{
\begin{tabular}{c|ccccc}
\hline
\multirow{2}{*}{\textbf{Dataset}} & \multicolumn{5}{c}{\textbf{Method}} \\
                                  & HLA  & TextHoaxer & LeapAttack & HQA-Attack & \textbf{LBA} \\ \hline
MR                                & 62.8 & 66.0       & 67.5       & 65.2       & \textbf{70.4} \\
IMDB                              & 64.4 & 88.6       & 86.9       & 88.3       & \textbf{90.6} \\ \hline
\end{tabular}}
\end{table}

\subsection{Ablation Study and Case Study\label{ablationstudy}}
We study LBA's performance with and without components of aggressiveness and quality on dataset MR against BERT under the tight query budget. To further validate the effectiveness of the aggressiveness component, we introduce the \textbf{label inversion hit rate} (LIHR), which measures the proportion of accepted samples that successfully induce label flipping.
The results are shown in Table \ref{tab:ablationstudy}.
The results show that the aggressiveness component indeed helps to improve LIHR without querying the target model. At the same time, another component enhances the quality of samples.
Overall, this demonstrates that both the aggressiveness and quality components play crucial roles in approximating the distribution of high-quality adversarial examples.
We also list some adversarial texts crafted by LBA, which are shown in Appendix \ref{b1}.
These adversarial texts further demonstrate that LBA can generate high-quality adversarial texts with low-query budgets.

\begin{table}[]
\caption{Ablation study of LBA's components on dataset MR against BERT. }
\label{tab:ablationstudy}
\centering
\resizebox{0.4\textwidth}{!}{
\begin{tabular}{c|cccc|c}
\hline
\textbf{Method}                     & \textbf{Pert.(\%)}     & \textbf{Sim.}           & \textbf{Ige.(\%)}     & \textbf{PPL}            & \textbf{LIHR (\%)}     \\ \hline
\multirow{2}{*}{Quality} & \multirow{2}{*}{16.57} & \multirow{2}{*}{0.8345} & \multirow{2}{*}{1.51} & \multirow{2}{*}{236.83} & \multirow{2}{*}{21.77} \\
                                    &                        &                         &                       &                         &                        \\ \hline
\multirow{2}{*}{Aggressiveness}        & \multirow{2}{*}{19.59} & \multirow{2}{*}{0.8276} & \multirow{2}{*}{1.87} & \multirow{2}{*}{263.44} & \multirow{2}{*}{30.56} \\
                                    &                        &                         &                       &                         &                        \\ \hline
\multirow{2}{*}{LBA}                & \multirow{2}{*}{14.03} & \multirow{2}{*}{0.8477} & \multirow{2}{*}{1.47} & \multirow{2}{*}{209.33} & \multirow{2}{*}{28.84} \\
                                    &                        &                         &                       &                         &                        \\ \hline
\end{tabular}
}
\end{table}

\section{Conclusion}
In this paper, we present LBA, a novel sampling-based hard-label attack that generates high-quality adversarial texts with low query budgets.
Our approach overcomes the limitations of existing methods by designing a sampling-based strategy to mitigate the explosion of the search space, thereby improving the query efficiency.
Through extensive experiments on six language models and four datasets, we demonstrate that LBA significantly outperforms existing hard-label attacks across all evaluation metrics. Furthermore, our results show that the adversarial texts generated by LBA are not only effective but also comprehensible, as evaluated by GPT-4o.

\section*{Acknowledgments}
This work was supported by the Key Project of the National Natural Science Foundation of China under grant no. 62536007, the National Natural Science Foundation of China under grant no. 62502454, the Zhejiang Province Science Foundation under grant no. LD24F020002, LQ24F020025, LZYQ25F020003, the Zhejiang Province’s 2025 "Leading Goose + X" Science and Technology Plan under grant no. 2025C02034, and the Jinhua Science and Technology Plan under Grant 2023-1-091.

\bibliographystyle{named}
\bibliography{ijcai26}

@article{alzantot2018generating,
	title={Generating Natural Language Adversarial Examples},
	author={Alzantot, Moustafa and Sharma, Yash and Elgohary, Ahmed and Ho, Bo-Jhang and Srivastava, B. Mani and Chang, Kai-Wei},
	journal={EMNLP},
	pages={2890--2896},
	year={2018}
}

@inproceedings{li2019textbugger,
	title={TextBugger: Generating Adversarial Text Against Real-world Applications},
	author={Li, Jinfeng and Ji, Shouling and Du, Tianyu and Li, Bo and Wang, Ting},
    booktitle={26th Annual Network and Distributed System Security Symposium, NDSS},
    year={2020}
}

@inproceedings{jin2020bert,
  title={Is bert really robust? a strong baseline for natural language attack on text classification and entailment},
  author={Jin, Di and Jin, Zhijing and Zhou, Joey Tianyi and Szolovits, Peter},
  booktitle={Proceedings of the AAAI conference on artificial intelligence},
  volume={34},
  number={05},
  pages={8018--8025},
  year={2020}
}

@article{zang2020word-level,
	title={Word-level Textual Adversarial Attacking as Combinatorial Optimization},
	author={Zang, Yuan and Qi, Fanchao and Yang, Chenghao and Liu, Zhiyuan and Zhang, Meng and Liu, Qun and Sun, Maosong},
    booktitle = "Proceedings of the 58th Annual Meeting of the Association for Computational Linguistics",
    year = "2020",
    publisher = "Association for Computational Linguistics",
    pages = "6066--6080",
}

@inproceedings{maheshwary2020,
  title={Generating natural language attacks in a hard label black box setting},
  author={Maheshwary, Rishabh and Maheshwary, Saket and Pudi, Vikram},
  booktitle={Proceedings of the AAAI Conference on Artificial Intelligence},
  volume={35},
  number={15},
  pages={13525--13533},
  year={2021}
}

@inproceedings{ye2022texthoaxer,
  title={TextHoaxer: budgeted hard-label adversarial attacks on text},
  author={Ye, Muchao and Miao, Chenglin and Wang, Ting and Ma, Fenglong},
  booktitle={Proceedings of the AAAI Conference on Artificial Intelligence},
  volume={36},
  number={4},
  pages={3877--3884},
  year={2022}
}

@inproceedings{ye2022leapattack,
  title={LeapAttack: Hard-Label Adversarial Attack on Text via Gradient-Based Optimization},
  author={Ye, Muchao and Chen, Jinghui and Miao, Chenglin and Wang, Ting and Ma, Fenglong},
  booktitle={Proceedings of the 28th ACM SIGKDD Conference on Knowledge Discovery and Data Mining},
  pages={2307--2315},
  year={2023}
}

@article{liu2024hqa,
  title={HQA-attack: toward high quality black-box hard-label adversarial attack on text},
  author={Liu, Han and Xu, Zhi and Zhang, Xiaotong and Zhang, Feng and Ma, Fenglong and Chen, Hongyang and Yu, Hong and Zhang, Xianchao},
  journal={Advances in Neural Information Processing Systems},
  volume={36:51347-51358},
  year={2024}
}

@article{chib1995understanding,
  title={Understanding the metropolis-hastings algorithm},
  author={Chib, Siddhartha and Greenberg, Edward},
  journal={The american statistician},
  volume={49},
  number={4},
  pages={327--335},
  year={1995},
  publisher={Taylor \& Francis}
}

@article{zhao2023survey,
  title={A survey of large language models},
  author={Zhao, Wayne Xin and Zhou, Kun and Li, Junyi and Tang, Tianyi and Wang, Xiaolei and Hou, Yupeng and Min, Yingqian and Zhang, Beichen and Zhang, Junjie and Dong, Zican and others},
  journal={arXiv preprint arXiv:2303.18223},
  year={2023}
}

@article{pang2005seeing,
	title={Seeing stars: exploiting class relationships for sentiment categorization with respect to rating scales},
	author={Pang, Bo and Lee, Lillian},
	journal={meeting of the association for computational linguistics},
	pages={115--124},
	year={2005}
}

@inproceedings{Zhang2015CharacterlevelCN,
author = {Zhang, Xiang and Zhao, Junbo and LeCun, Yann},
title = {Character-level convolutional networks for text classification},
year = {2015},
address = {Cambridge, MA, USA},
booktitle = {Proceedings of the 29th International Conference on Neural Information Processing Systems - Volume 1},
pages = {649–657},
series = {NIPS'15}
}

@article{maas2011learning,
	title={Learning word vectors for sentiment analysis},
	author={Maas, L. Andrew and Daly, E. Raymond and Pham, T. Peter and Huang, Dan and Ng, Y. Andrew and Potts, Christopher},
	journal={ACL},
	pages={142--150},
	year={2011}
}

@article{kim2014convolutional,
	title={Convolutional Neural Networks for Sentence Classification},
	author={Kim, Yoon},
	journal={EMNLP},
	pages={1746--1751},
	year={2014}
}

@article{s1997long,
	title={Long short-term memory},
	author={Hochreiter, Sepp and Schmidhuber, Jurgen},
    year = {1997},
    issue_date = {November 15, 1997},
    publisher = {MIT Press},
    address = {Cambridge, MA, USA},
    volume = {9},
    number = {8},
    issn = {0899-7667},
    journal = {Neural Comput.},
    pages = {1735–1780},
    numpages = {46}
}

@inproceedings{devlin2019bert,
  title={Bert: Pre-training of deep bidirectional transformers for language understanding},
  author={Devlin, Jacob and Chang, Ming-Wei and Lee, Kenton and Toutanova, Kristina},
  booktitle={Proceedings of the 2019 conference of the North American chapter of the association for computational linguistics: human language technologies, volume 1 (long and short papers)},
  pages={4171--4186},
  year={2019}
}

@article{touvron2023llama,
  title={Llama: Open and efficient foundation language models},
  author={Touvron, Hugo and Lavril, Thibaut and Izacard, Gautier and Martinet, Xavier and Lachaux, Marie-Anne and Lacroix, Timoth{\'e}e and Rozi{\`e}re, Baptiste and Goyal, Naman and Hambro, Eric and Azhar, Faisal and others},
  journal={arXiv preprint arXiv:2302.13971},
  year={2023}
}

@inproceedings{vicuna2023,
  title={GLM: General Language Model Pretraining with Autoregressive Blank Infilling},
  author={Du, Zhengxiao and Qian, Yujie and Liu, Xiao and Ding, Ming and Qiu, Jiezhong and Yang, Zhilin and Tang, Jie},
  booktitle={Proceedings of the 60th Annual Meeting of the Association for Computational Linguistics (Volume 1: Long Papers)},
  pages={320--335},
  year={2022}
}

@article{openai2023gpt,
  title={GPT-4 technical report},
  author={OpenAI, R},
  journal={arXiv},
  pages={2303-08774},
  year={2023}
}

@article{radford2019language,
  title={Language models are unsupervised multitask learners},
  author={Radford, Alec and Wu, Jeffrey and Child, Rewon and Luan, David and Amodei, Dario and Sutskever, Ilya and others},
  journal={OpenAI blog},
  volume={1},
  number={8},
  pages={9},
  year={2019}
}

@inproceedings{li2021contextualized,
  title={Contextualized Perturbation for Textual Adversarial Attack},
  author={Li, Dianqi and Zhang, Yizhe and Peng, Hao and Chen, Liqun and Brockett, Chris and Sun, Ming-Ting and Dolan, William B},
  booktitle={Proceedings of the 2021 Conference of the North American Chapter of the Association for Computational Linguistics: Human Language Technologies},
  pages={5053--5069},
  year={2021}
}

@article{peng2023textcheater,
  title={TextCheater: A Query-Efficient Textual Adversarial Attack in the Hard-Label Setting},
  author={Peng, Hao and Guo, Shixin and Zhao, Dandan and Zhang, Xuhong and Han, Jianmin and Ji, Shouling and Yang, Xing and Zhong, Ming},
  journal={IEEE Transactions on Dependable and Secure Computing},
  number={01},
  pages={1--16},
  year={2023},
  publisher={IEEE Computer Society}
}

@inproceedings{cer2018universal,
  title={Universal sentence encoder for English},
  author={Cer, Daniel and Yang, Yinfei and Kong, Sheng-yi and Hua, Nan and Limtiaco, Nicole and John, Rhomni St and Constant, Noah and Guajardo-Cespedes, Mario and Yuan, Steve and Tar, Chris and others},
  booktitle={Proceedings of the 2018 conference on empirical methods in natural language processing: system demonstrations},
  pages={169--174},
  year={2018}
}

@inproceedings{Sze2014intriguing,
title = {Intriguing properties of neural networks},
author = {Christian Szegedy and Wojciech Zaremba and Ilya Sutskever and Joan Bruna and Dumitru Erhan and Ian Goodfellow and Rob Fergus},
year = {2014},
booktitle = {International Conference on Learning Representations (ICLR)},
}

@inproceedings{GoodfellowSS14,
  author       = {Ian J. Goodfellow and
                  Jonathon Shlens and
                  Christian Szegedy},
  title        = {Explaining and Harnessing Adversarial Examples},
  booktitle    = {International Conference on Learning Representations (ICLR)},
  year         = {2015}
}

@inproceedings{morris2020textattack,
  title={TextAttack: A Framework for Adversarial Attacks, Data Augmentation, and Adversarial Training in NLP},
  author={Morris, John and Lifland, Eli and Yoo, Jin Yong and Grigsby, Jake and Jin, Di and Qi, Yanjun},
  booktitle={Proceedings of the 2020 Conference on Empirical Methods in Natural Language Processing: System Demonstrations},
  pages={119--126},
  year={2020}
}

@article{linyang2020bert,
	title={BERT ATTACK: Adversarial Attack Against BERT Using BERT},
	author={Linyang, Li and Ruotian, Ma and Qipeng, Guo and Xiangyang, Xue and Xipeng, Qiu},
	journal={EMNLP 2020},
	pages={6193--6202},
	year={2020}
}

@inproceedings{maheshwary2021strong,
  title={A Strong Baseline for Query Efficient Attacks in a Black Box Setting},
  author={Maheshwary, Rishabh and Maheshwary, Saket and Pudi, Vikram},
  booktitle={Proceedings of the 2021 Conference on Empirical Methods in Natural Language Processing},
  pages={8396--8409},
  year={2021}
}

@inproceedings{gao2024semantic,
  title={Semantic-Preserving Adversarial Example Attack against BERT},
  author={Gao, Chongyang and Gu, Kang and Vosoughi, Soroush and Mehnaz, Shagufta},
  booktitle={Proceedings of the 4th Workshop on Trustworthy Natural Language Processing (TrustNLP 2024)},
  pages={202--207},
  year={2024}
}

@inproceedings{garg2020bae,
  title={BAE: BERT-based Adversarial Examples for Text Classification},
  author={Garg, Siddhant and Ramakrishnan, Goutham},
  booktitle={Proceedings of the 2020 Conference on Empirical Methods in Natural Language Processing (EMNLP)},
  pages={6174--6181},
  year={2020}
}

@inproceedings{hu-etal-2022-deep,
    title = "{DEEP}: {DE}noising Entity Pre-training for Neural Machine Translation",
    author = "Hu, Junjie  and
      Hayashi, Hiroaki  and
      Cho, Kyunghyun  and
      Neubig, Graham",
    editor = "Muresan, Smaranda  and
      Nakov, Preslav  and
      Villavicencio, Aline",
    booktitle = "Proceedings of the 60th Annual Meeting of the Association for Computational Linguistics (Volume 1: Long Papers)",
    year = "2022",
    pages = "1753--1766",
}

@inproceedings{zhangicml23,
author = {Zhang, Biao and Haddow, Barry and Birch, Alexandra},
title = {Prompting large language model for machine translation: a case study},
year = {2023},
booktitle = {Proceedings of the 40th International Conference on Machine Learning},
pages={41092--41110},
series = {ICML'23}
}

@article{brown2020language,
  title={Language models are few-shot learners},
  author={Brown, Tom and Mann, Benjamin and Ryder, Nick and Subbiah, Melanie and Kaplan, Jared D and Dhariwal, Prafulla and Neelakantan, Arvind and Shyam, Pranav and Sastry, Girish and Askell, Amanda and others},
  journal={Advances in neural information processing systems},
  volume={33},
  pages={1877--1901},
  year={2020}
}

@article{LIU2024106461,
title = {HyGloadAttack: Hard-label black-box textual adversarial attacks via hybrid optimization},
journal = {Neural Networks},
volume = {178},
pages = {106461},
year = {2024},
author = {Zhaorong Liu and Xi Xiong and Yuanyuan Li and Yan Yu and Jiazhong Lu and Shuai Zhang and Fei Xiong},
}

\clearpage

\appendix
\renewcommand{\thefigure}{\arabic{figure}}
\renewcommand{\thetable}{\arabic{table}}
\setcounter{figure}{0}
\setcounter{table}{0}
\section{Additional Experiment Result}
\subsection{Attack Performance against WordLSTM and WordCNN}
\label{a1}
We present attack results against WordCNN and WordLSTM under the given query budgets. As shown in Tables \ref{tab:attack-on-lstm} and \ref{tab:attack-on-cnn}, LBA consistently achieves the highest similarity and fluency across all datasets and query budgets. These results highlight LBA's capability to generate high-quality adversarial texts under low query budgets.

\subsection{Distribution of Perturbation Rate and Semantic Similarity for Adversarial Examples}
\label{a2}
\begin{figure*}[]
    \centering
    % 第一行
    \subfloat[MR (Pert.)\label{fig:image1}]{
        \includegraphics[width=0.22\textwidth]{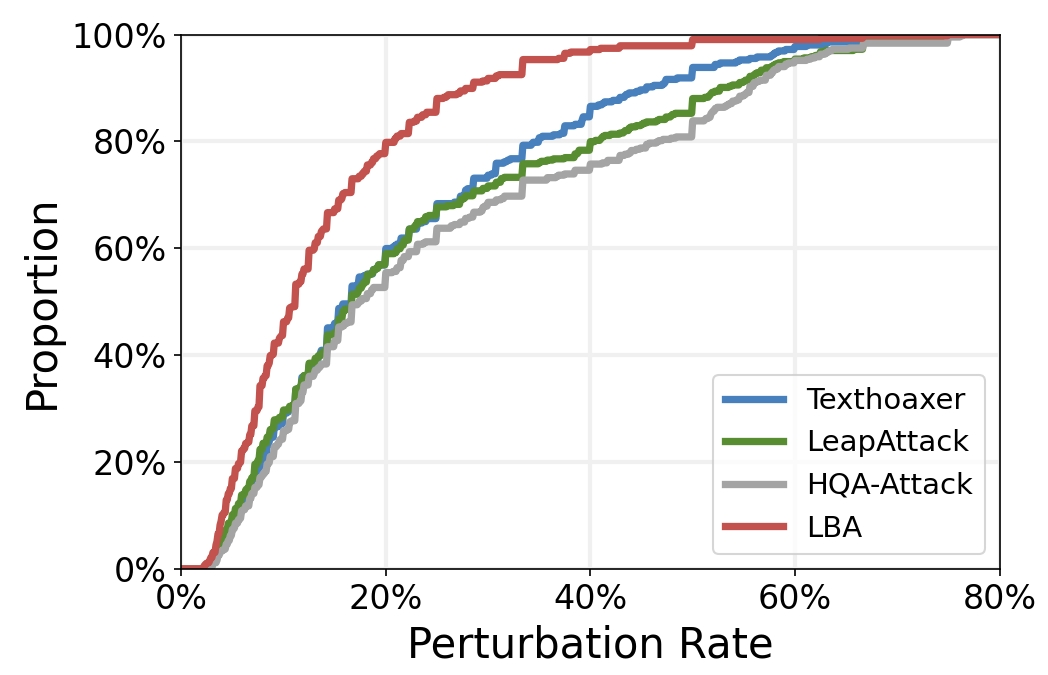}
    }
    \hspace{0.01\textwidth} % 子图之间的间距
    \subfloat[AG (Pert.)\label{fig:image2}]{
        \includegraphics[width=0.22\textwidth]{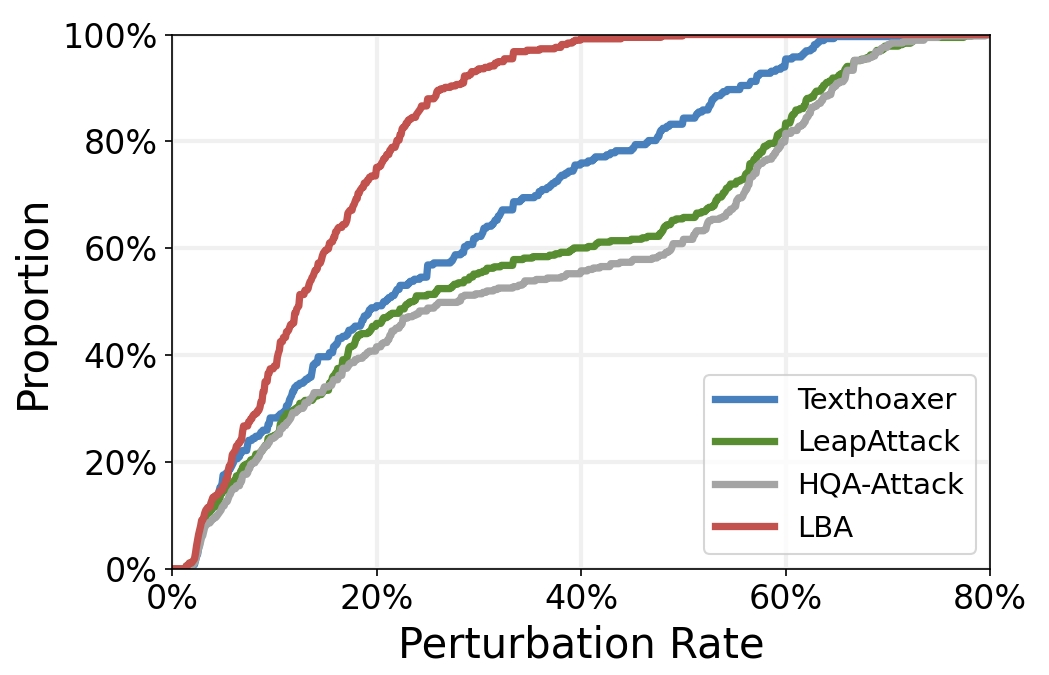}
    }
    \hspace{0.01\textwidth}
    \subfloat[YELP (Pert.)\label{fig:image3}]{
        \includegraphics[width=0.22\textwidth]{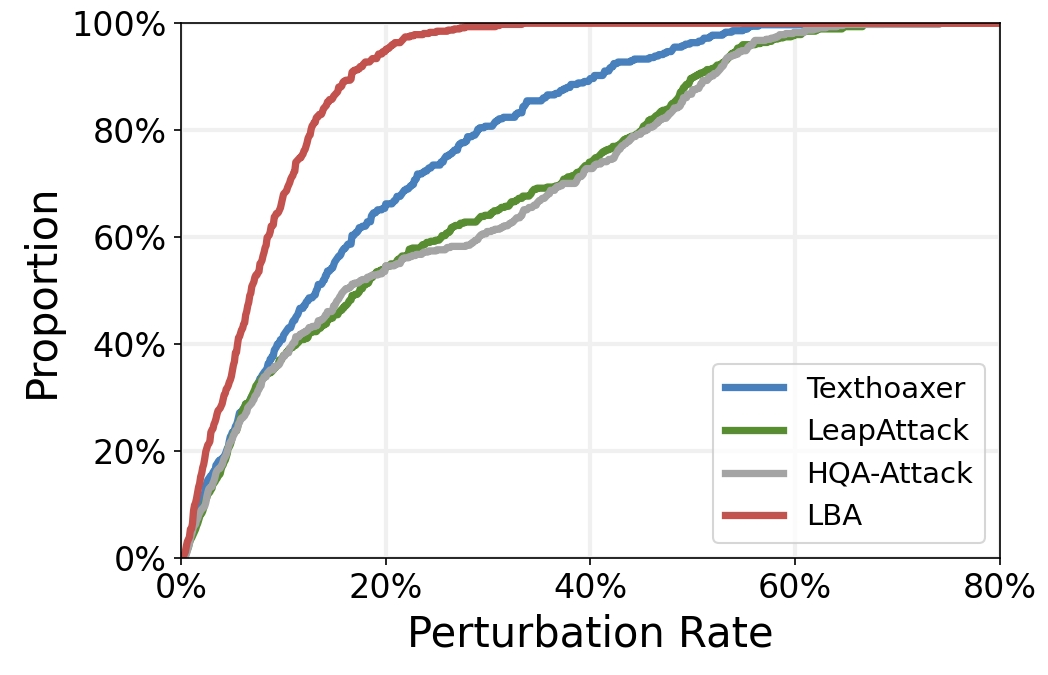}
    }
    \hspace{0.01\textwidth}
    \subfloat[IMDB (Pert.)\label{fig:image4}]{
        \includegraphics[width=0.22\textwidth]{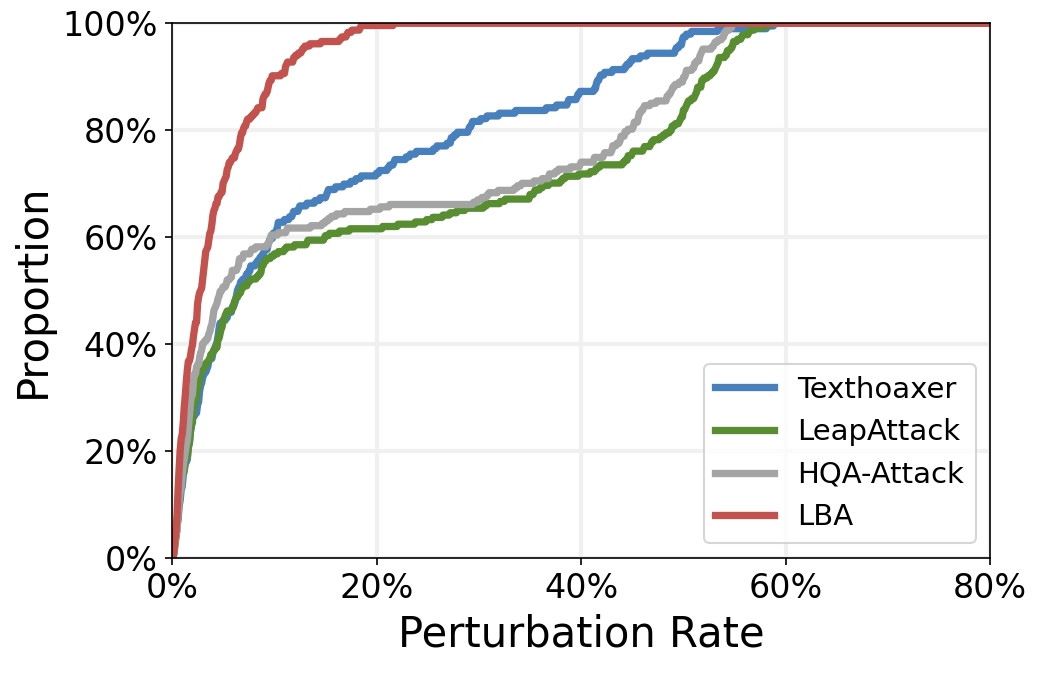}
    }

    \subfloat[MR (Sim.)\label{fig:image5}]{
        \includegraphics[width=0.22\textwidth]{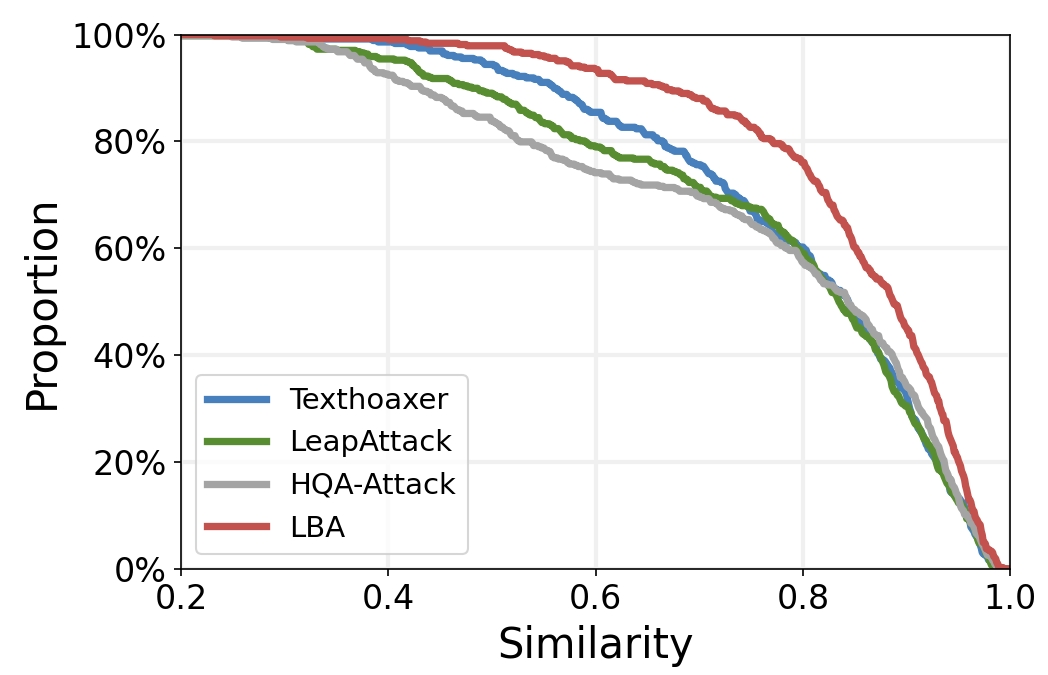}
    }
    \hspace{0.01\textwidth}
    \subfloat[AG (Sim.)\label{fig:image6}]{
        \includegraphics[width=0.22\textwidth]{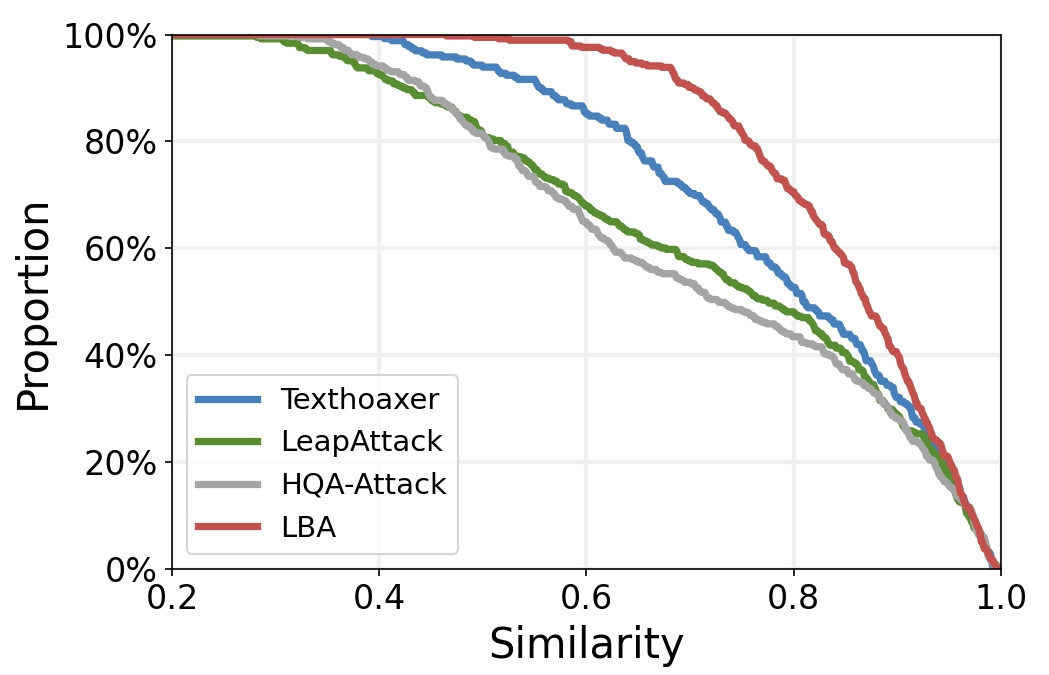}
    }
    \hspace{0.01\textwidth}
    \subfloat[YELP (Sim.)\label{fig:image7}]{
        \includegraphics[width=0.22\textwidth]{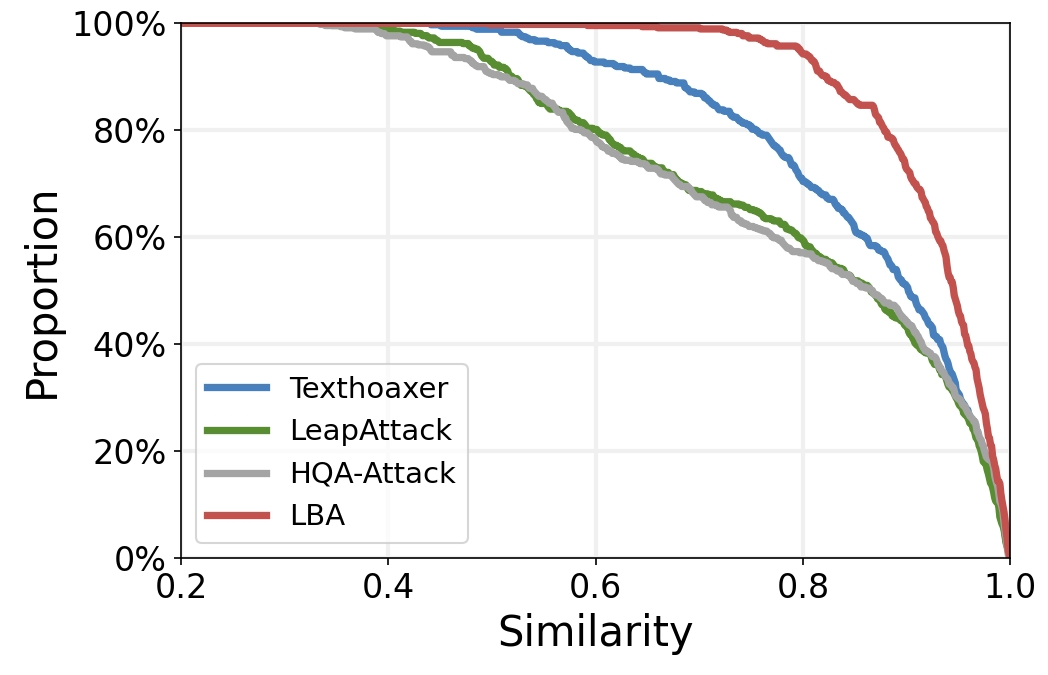}
    }
    \hspace{0.01\textwidth}
    \subfloat[IMDB (Sim.)\label{fig:image8}]{
        \includegraphics[width=0.22\textwidth]{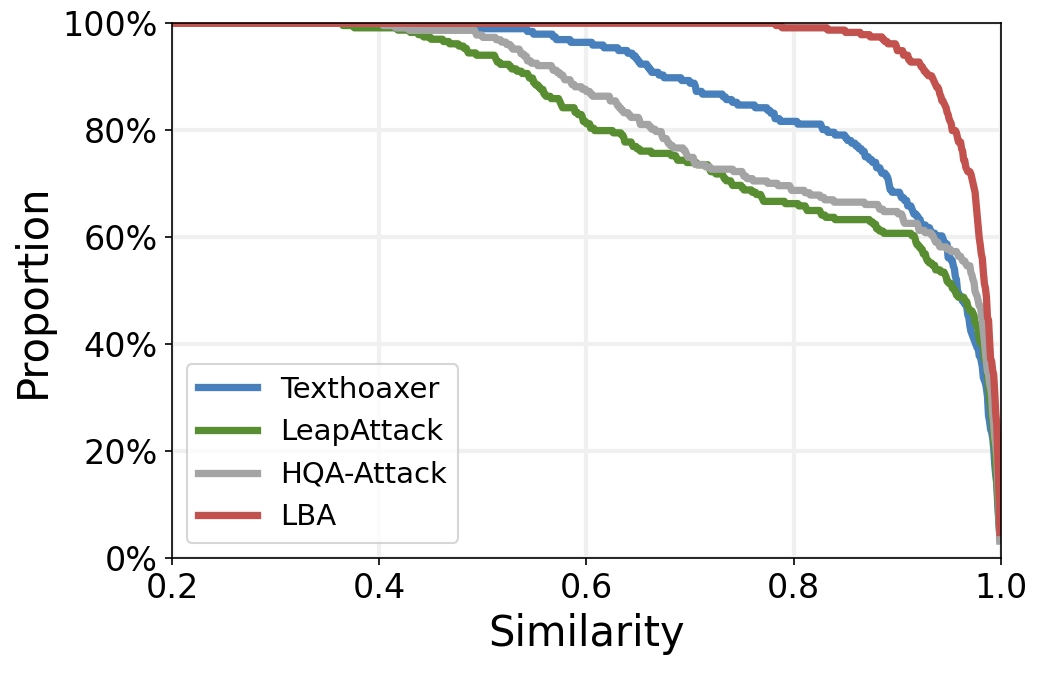}
    }
\caption{
Proportion of adversarial examples achieving at most a given perturbation rate (a–d) and at least a given semantic similarity (e–h) against BERT under tiny query budgets.
}
\label{asr-sim}
\end{figure*}

To further analyze the distribution of generated samples in terms of similarity, we compare the cumulative distributions of perturbation rate and semantic similarity in Figure \ref{asr-sim}. Across all datasets, LBA's curves consistently dominate other baselines, rising faster in the proportion of low perturbation rate samples and staying higher for semantic similarity. These results highlight LBA's advantage in generating a higher proportion of low-perturbation, high-semantic similarity adversarial texts under tiny query budgets.

\section{Case Study}
\label{b1}
We list some adversarial texts crafted by LeapAttack, HQA-Attack and LBA under the tight query budget, as illustrated in Table \ref{tab:casestudy}. For instance, the adversarial example generated by LBA demonstrates superior similarity and fluency compared to those generated by other hard-label attacks, successfully fooling LLaMA-7b-chat by substituting "notch" and ``prompt” with ``buttocks” and ``expeditious,” respectively.
These two attributes of adversarial texts are crucial for aligning with the definition of adversarial texts, which requires them to be imperceptible to humans yet capable of causing target models to make incorrect predictions.
\begin{table}[h]
\caption{Parameter tuning results for the quality component weights $\lambda_1$, $\lambda_2$, $\lambda_3$, and $\lambda_4$ in the target function, evaluated on BERT-MR under the tight budget setting.}
\centering
\label{Parameters_tuning}
\resizebox{0.38\textwidth}{!}{
\begin{tabular}{cccc|cc}
\hline
\multicolumn{4}{c|}{\textbf{Weight Parameters}}               & \multicolumn{2}{c}{\textbf{Evaluation Metrics}} \\
$\lambda_1$   & $\lambda_2$   & $\lambda_3$   & $\lambda_4$   & Pert.(\%)               & PPL                    \\ \hline
0.25          & 0.25          & 0.25          & 0.25          & 14.23                  & 228.41                 \\
0.40          & 0.40          & 0.40          & 0.40          & 13.67                  & 251.83                 \\
0.10          & 0.10          & 0.40          & 0.40          & 15.52                  & 189.26                 \\
0.15          & 0.60          & 0.15          & 0.10          & 13.89                  & 219.54                 \\
\textbf{0.25} & \textbf{0.50} & \textbf{0.15} & \textbf{0.10} & \textbf{12.41}         & \textbf{207.07}        \\
0.35          & 0.40          & 0.15          & 0.10          & 13.46                  & 218.72                 \\ \hline
\end{tabular}}
\end{table}
\section{Justification of Weight Parameters}
The weights of the quality component satisfy $\lambda_1+\lambda_2+\lambda_3+\lambda_4=1$, where $\lambda_1, \lambda_2$ correspond to semantic similarity and $\lambda_3, \lambda_4$ correspond to fluency. We determined these weights through a two-stage sensitivity analysis. First, we searched for the optimal allocation between the similarity group ($\lambda_1+\lambda_2$) and the fluency group ($\lambda_3+\lambda_4$). Then, we tuned the ratio within each group using adaptive step sizes. As shown in Table~\ref{Parameters_tuning}, the uniform allocation (Row 1) yields suboptimal performance on both metrics. Biasing toward similarity (Row 2) reduces perturbation rate but significantly degrades fluency, while biasing toward fluency (Row 3) improves PPL at the cost of much higher perturbation. Rows 4 and 6 represent configurations near the optimum with slight variations. The optimal setting $\lambda_1=0.25, \lambda_2=0.50, \lambda_3=0.15, \lambda_4=0.10$ (Row 5, bolded) achieves the best trade-off, with the lowest perturbation rate and a competitive PPL.

\begin{table*}[!t]
\centering
\caption{Comparison of Pert., Sim., Ige. and
PPL with given query budgets when attacking against WordLSTM.}
\label{tab:attack-on-lstm}
\resizebox{1\textwidth}{!}{
\begin{tabular}{cc|ccccc|ccccc|ccccc}
\hline 
\multirow{2}{*}{\textbf{Dataset}}          & \multirow{2}{*}{\textbf{Method}} & \multicolumn{5}{c|}{\textbf{Tiny Budget}}                                                           & \multicolumn{5}{c|}{\textbf{Tight Budget}}                                                          & \multicolumn{5}{c}{\textbf{Moderate Budget}}                                                        \\ \cline{3-17} 
                                           &                                  & \textbf{ASR.(\%)}      & \textbf{Pert.(\%)} & \textbf{Sim.}   & \textbf{Ige.(\%)} & \textbf{PPL}    & \textbf{ASR.(\%)}      & \textbf{Pert.(\%)} & \textbf{Sim.}   & \textbf{Ige.(\%)} & \textbf{PPL}    & \textbf{ASR.(\%)}      & \textbf{Pert.(\%)} & \textbf{Sim.}   & \textbf{Ige.(\%)} & \textbf{PPL}    \\ \hline
\multicolumn{1}{c|}{\multirow{7}{*}{MR}}   & \textit{Textbugger}              & \underline{47.32}        & 11.10              & 0.9242          & 4.66              & 205.34          & \underline{76.92}       & 12.19              & 0.9083          & 4.79              & 202.50          & 80.19                  & 12.65              & 0.9063          & 4.87              & 210.61          \\
\multicolumn{1}{c|}{}                      & \textit{Textfooler}              & \underline{27.04}        & 11.86              & 0.8800          & 0.65              & 202.84          & 79.02                  & 11.79              & 0.8826          & 0.82              & 164.99          & 95.57                  & 12.81              & 0.8722          & 0.98              & 171.22          \\ \cline{3-17} 
\multicolumn{1}{c|}{}                      & HLA                              & \multirow{5}{*}{94.41} & 25.11              & 0.7305          & 1.88              & 326.14          & \multirow{5}{*}{97.20} & 21.66              & 0.7885          & 1.67              & 260.04          & \multirow{5}{*}{97.44} & 20.30              & 0.7947          & 1.66              & 246.09          \\
\multicolumn{1}{c|}{}                      & Texthoaxer                       &                        & 19.99              & 0.8042          & 1.72              & 260.48          &                        & 14.19              & 0.8598          & 1.14              & 199.35          &                        & 13.25              & 0.8740          & 1.10              & 187.71          \\
\multicolumn{1}{c|}{}                      & LeapAttack                       &                        & 19.79              & 0.8061          & 1.79              & 260.24          &                        & 13.58              & 0.8675          & 1.09              & 191.82          &                        & 12.30              & 0.8811          & 0.91              & 182.48          \\
\multicolumn{1}{c|}{}                      & HQA-Attack                       &                        & 20.50              & 0.8089          & 1.44              & 256.82          &                        & 12.95              & 0.8859          & 0.59              & 178.50          &                        & 12.36              & 0.8924          & 0.75              & 177.39          \\
\multicolumn{1}{c|}{}                      & \textbf{LBA}                     &                        & \textbf{12.04}     & \textbf{0.8624} & \textbf{0.97}     & \textbf{180.09} &                        & \textbf{10.84}     & \textbf{0.8874} & \textbf{0.47}     & \textbf{172.37} &                        & \textbf{10.32}     & \textbf{0.8932} & \textbf{0.67}     & \textbf{167.30} \\ \hline
\multicolumn{1}{c|}{\multirow{7}{*}{AG}}   & \textit{Textbugger}              & \underline{33.33}        & 10.22              & 0.9267          & 2.80              & 248.23          & 70.18                  & 16.99              & 0.8942          & 4.46              & 293.46          & 74.56                  & 17.96              & 0.8889          & 4.78              & 313.42          \\
\multicolumn{1}{c|}{}                      & \textit{Textfooler}              & \underline{22.81}        & 7.16               & 0.9368          & 0.47              & 187.86          & \underline{54.61}        & 11.30              & 0.9035          & 0.93              & 232.56          & 75.00                  & 14.76              & 0.8795          & 1.35              & 267.66          \\ \cline{3-17} 
\multicolumn{1}{c|}{}                      & HLA                              & \multirow{5}{*}{77.41} & 38.05              & 0.6748          & 2.54              & 642.59          & \multirow{5}{*}{85.31} & 33.73              & 0.7263          & 2.37              & 593.53          & \multirow{5}{*}{87.50} & 27.63              & 0.7926          & 2.44              & 464.66          \\
\multicolumn{1}{c|}{}                      & Texthoaxer                       &                        & 27.24              & 0.7555          & 1.87              & 544.73          &                        & 19.64              & 0.8344          & 1.35              & 380.10          &                        & 19.20              & 0.8381          & 0.96              & 372.26          \\
\multicolumn{1}{c|}{}                      & LeapAttack                       &                        & 31.96              & 0.7312          & 2.15              & 564.64          &                        & 20.69              & 0.8323          & 1.28              & 392.49          &                        & 18.02              & 0.8519          & 1.22              & 351.04          \\
\multicolumn{1}{c|}{}                      & HQA-Attack                       &                        & 31.78              & 0.7381          & 2.15              & 584.01          &                        & 20.36              & 0.8333          & 1.43              & 391.30          &                        & 16.83              & 0.8684          & 0.90              & 330.68          \\
\multicolumn{1}{c|}{}                      & \textbf{LBA}                     &                        & \textbf{15.98}     & \textbf{0.8434} & \textbf{1.01}     & \textbf{328.57} &                        & \textbf{15.00}     & \textbf{0.8525} & \textbf{0.95}     & \textbf{318.71} &                        & \textbf{14.93}     & \textbf{0.8694} & \textbf{0.61}     & \textbf{312.95} \\ \hline
\multicolumn{1}{c|}{\multirow{7}{*}{YELP}} & \textit{Textbugger}              & \underline{52.81}        & 8.60               & 0.9406          & 2.79              & 112.16          & 86.58                  & 7.56               & 0.9503          & 2.47              & 90.75           & 92.86                  & 7.04               & 0.9544          & 2.23              & 85.47           \\
\multicolumn{1}{c|}{}                      & \textit{Textfooler}              & \underline{43.72}        & 6.27               & 0.9530          & 0.52              & 94.99           & 85.71                  & 5.63               & 0.9605          & 0.56              & 81.61           & 96.75                  & 5.31               & 0.9635          & 0.54              & 79.94           \\ \cline{3-17} 
\multicolumn{1}{c|}{}                      & HLA                              & \multirow{5}{*}{97.84} & 29.69              & 0.7292          & 2.65              & 207.60          & \multirow{5}{*}{99.13} & 20.86              & 0.8048          & 2.35              & 274.78          & \multirow{5}{*}{99.78} & 15.67              & 0.8795          & 1.86              & 185.08          \\
\multicolumn{1}{c|}{}                      & Texthoaxer                       &                        & 14.86              & 0.8807          & 1.72              & 155.96          &                        & 10.79              & 0.9194          & 1.22              & 133.08          &                        & 8.07               & 0.9416          & 0.90              & 98.37           \\
\multicolumn{1}{c|}{}                      & LeapAttack                       &                        & 17.49              & 0.8528          & 2.07              & 205.68          &                        & 10.19              & 0.9198          & 1.14              & 130.35          &                        & 7.25               & 0.9480          & 1.10              & 97.71           \\
\multicolumn{1}{c|}{}                      & HQA-Attack                       &                        & 16.53              & 0.8649          & 1.96              & 194.26          &                        & 9.76               & 0.9279          & 1.02              & 128.73          &                        & 6.91               & 0.9526          & 0.69              & 97.05           \\
\multicolumn{1}{c|}{}                      & \textbf{LBA}                     &                        & \textbf{5.82}      & \textbf{0.9479} & \textbf{0.65}     & \textbf{84.30}  &                        & \textbf{5.20}      & \textbf{0.9551} & \textbf{0.51}     & \textbf{80.54}  &                        & \textbf{4.98}      & \textbf{0.9563} & \textbf{0.45}     & \textbf{79.16}  \\ \hline
\multicolumn{1}{c|}{\multirow{7}{*}{IMDB}} & \textit{Textbugger}              & \underline{70.73}        & 4.32               & 0.9752          & 1.75              & 88.11           & 85.37                  & 4.23               & 0.9769          & 1.70              & 85.59           & 87.40                  & 4.47               & 0.9756          & 1.66              & 87.46           \\
\multicolumn{1}{c|}{}                      & \textit{Textfooler}              & \underline{66.26}        & 2.73               & 0.9845          & 0.22              & 76.68           & 87.80                  & 2.70               & 0.9851          & 0.27              & 76.01           & 88.62                  & 2.74               & 0.9849          & 0.27              & 76.78           \\ \cline{3-17} 
\multicolumn{1}{c|}{}                      & HLA                              & \multirow{5}{*}{98.78} & 18.20              & 0.8517          & 2.21              & 221.96          & \multirow{5}{*}{99.19} & 15.90              & 0.8895          & 1.86              & 183.14          & \multirow{5}{*}{99.19} & 14.31              & 0.9005          & 1.67              & 180.02          \\
\multicolumn{1}{c|}{}                      & Texthoaxer                       &                        & 8.97               & 0.9433          & 0.98              & 128.54          &                        & 4.87               & 0.9722          & 0.55              & 91.41           &                        & 3.40               & 0.9809          & 0.30              & 83.18           \\
\multicolumn{1}{c|}{}                      & LeapAttack                       &                        & 11.08              & 0.9167          & 1.29              & 143.75          &                        & 4.54               & 0.9708          & 0.52              & 93.19           &                        & 2.73               & 0.9841          & 0.31              & 79.70           \\
\multicolumn{1}{c|}{}                      & HQA-Attack                       &                        & 10.20              & 0.9285          & 1.18              & 137.40          &                        & 5.17               & 0.9657          & 0.56              & 98.55           &                        & 2.17               & 0.9893          & 0.23              & 75.10           \\
\multicolumn{1}{c|}{}                      & \textbf{LBA}                     &                        & \textbf{1.98}      & \textbf{0.9877} & \textbf{0.15}     & \textbf{74.92}  &                        & \textbf{1.93}      & \textbf{0.9882} & \textbf{0.17}     & \textbf{75.31}  &                        & \textbf{1.85}      & \textbf{0.9919} & \textbf{0.16}     & \textbf{74.34}  \\ \hline 
\end{tabular}
}
\end{table*}
\begin{table*}[!t]
\centering
\caption{Comparison of Pert., Sim., Ige. and PPL with given query budgets when attacking against WordCNN.}
\label{tab:attack-on-cnn}
\resizebox{1\textwidth}{!}{
\begin{tabular}{c|c|ccccc|ccccc|ccccc}
\hline
\multirow{2}{*}{\textbf{Dataset}} & \multirow{2}{*}{\textbf{Method}} & \multicolumn{5}{c|}{\textbf{Ting Budget}}                                                                    & \multicolumn{5}{c|}{\textbf{Tight Budget}}                                                                    & \multicolumn{5}{c}{\textbf{Moderate Budget}}                                                                  \\ \cline{3-17} 
                                  &                                  & \textbf{ASR.(\%)}               & \textbf{Pert.(\%)} & \textbf{Sim.}   & \textbf{Ige.(\%)} & \textbf{PPL}    & \textbf{ASR.(\%)}                & \textbf{Pert.(\%)} & \textbf{Sim.}   & \textbf{Ige.(\%)} & \textbf{PPL}    & \textbf{ASR.(\%)}                & \textbf{Pert.(\%)} & \textbf{Sim.}   & \textbf{Ige.(\%)} & \textbf{PPL}    \\ \hline
\multirow{7}{*}{MR}               & \textit{Textbugger}              & \underline{42.46}                 & 11.83              & 0.9167          & 5.03              & 215.78          & \underline{75.17}                  & 13.47              & 0.9058          & 5.50              & 213.54          & 78.42                            & 13.46              & 0.9052          & 5.36              & 215.62          \\
                                  & \textit{Textfooler}              & \underline{26.22}                 & 11.45              & 0.8780          & 1.20              & 192.05          & \underline{74.71}                  & 12.93              & 0.8741          & 1.11              & 177.51          & 96.06                            & 13.62              & 0.8689          & 1.36              & 181.16          \\ \cline{3-17} 
                                  & HLA                              & \multirow{5}{*}{95.13}          & 24.58              & 0.7386          & 2.31              & 316.31          & \multirow{5}{*}{96.98}           & 21.41              & 0.7913          & 1.57              & 253.09          & \multirow{5}{*}{98.14}           & 20.17              & 0.8002          & 1.62              & 234.53          \\
                                  & Texthoaxer                       &                                 & 19.27              & 0.8130          & 1.88              & 268.73          &                                  & 15.38              & 0.8535          & 1.74              & 211.34          &                                  & 14.30              & 0.8660          & 1.44              & 202.55          \\
                                  & LeapAttack                       &                                 & 20.92              & 0.7959          & 2.11              & 262.79          &                                  & 15.03              & 0.8531          & 1.53              & 208.60          &                                  & 13.17              & 0.8747          & 1.55              & 193.90          \\
                                  & HQA-Attack                       &                                 & 20.47              & 0.8067          & 2.10              & 269.86          &                                  & 15.34              & 0.8639          & 1.45              & 200.73          &                                  & 13.67              & 0.8821          & 1.47              & 187.12          \\
                                  & \textbf{LBA}                     &                                 & \textbf{13.33}     & \textbf{0.8589} & \textbf{1.61}     & \textbf{196.23} &                                  & \textbf{12.04}     & \textbf{0.8741} & \textbf{0.94}     & \textbf{185.76} &                                  & \textbf{11.30}     & \textbf{0.8879} & \textbf{1.20}     & \textbf{180.10} \\ \hline
\multirow{7}{*}{AG}               & \textit{Textbugger}              & \underline{38.99}                 & 10.74              & 0.9273          & 3.21              & 250.65          & 82.38                            & 16.81              & 0.8908          & 5.01              & 299.17          & 85.02                            & 17.62              & 0.8879          & 5.13              & 302.82          \\
                                  & \textit{Textfooler}              & \underline{22.47}                 & 6.92               & 0.9413          & 0.88              & 209.66          & 69.16                            & 12.45              & 0.8964          & 1.16              & 246.26          & 89.21                            & 14.50              & 0.8814          & 1.24              & 267.55          \\ \cline{3-17} 
                                  & HLA                              & \multirow{5}{*}{81.94}          & 37.80              & 0.6990          & 2.26              & 632.33          & \multirow{5}{*}{86.78}           & 30.90              & 0.7345          & 2.21              & 537.90          & \multirow{5}{*}{88.33}           & 25.81              & 0.8113          & 1.90              & 455.48          \\
                                  & Texthoaxer                       &                                 & 24.76              & 0.8024          & 1.93              & 474.05          &                                  & 17.25              & 0.8569          & 1.50              & 346.97          &                                  & 16.53              & 0.8677          & 1.25              & 334.12          \\
                                  & LeapAttack                       &                                 & 30.30              & 0.7544          & 2.23              & 557.03          &                                  & 18.21              & 0.8540          & 1.36              & 358.71          &                                  & 14.36              & 0.8877          & 1.13              & 303.14          \\
                                  & HQA-Attack                       &                                 & 29.41              & 0.7600          & 2.29              & 535.74          &                                  & 17.58              & 0.8656          & 1.49              & 348.67          &                                  & 13.91              & 0.8930          & 1.28              & 285.83          \\
                                  & \textbf{LBA}                     &                                 & \textbf{13.93}     & \textbf{0.8703} & \textbf{1.11}     & \textbf{298.02} &                                  & \textbf{13.04}     & \textbf{0.8723} & \textbf{0.68}     & \textbf{287.02} &                                  & \textbf{12.64}     & \textbf{0.8972} & \textbf{0.71}     & \textbf{276.90} \\ \hline
\multirow{7}{*}{YELP}             & \textit{Textbugger}              & \underline{45.69}                 & 8.36               & 0.9424          & 2.80              & 113.76          & 81.47                            & 8.29               & 0.9455          & 2.94              & 98.27           & 90.09                            & 8.05               & 0.9477          & 2.88              & 89.66           \\
                                  & \textit{Textfooler}              & \underline{37.07}                 & 6.83               & 0.9463          & 0.83              & 99.61           & 82.54                            & 6.41               & 0.9513          & 0.83              & 83.42           & 95.69                            & 6.03               & 0.9555          & 0.77              & 81.75           \\ \cline{3-17} 
                                  & HLA                              & \multirow{5}{*}{96.77}          & 32.81              & 0.7377          & 1.89              & 368.86          & \multirow{5}{*}{98.92}           & 23.39              & 0.8081          & 1.88              & 261.08          & \multirow{5}{*}{99.35}           & 18.18              & 0.8462          & 1.32              & 212.39          \\
                                  & Texthoaxer                       &                                 & 15.39              & 0.8780          & 1.75              & 151.46          &                                  & 11.92              & 0.9062          & 1.53              & 137.03          &                                  & 9.88               & 0.9283          & 1.17              & 109.21          \\
                                  & LeapAttack                       &                                 & 20.73              & 0.8241          & 2.51              & 233.10          &                                  & 12.33              & 0.9051          & 1.44              & 152.25          &                                  & 8.75               & 0.9360          & 1.07              & 110.67          \\
                                  & HQA-Attack                       &                                 & 19.58              & 0.8330          & 2.43              & 209.23          &                                  & 12.12              & 0.9082          & 1.47              & 147.33          &                                  & 8.21               & 0.9452          & 1.02              & 107.41          \\
                                  & \textbf{LBA}                     &                                 & \textbf{6.82}      & \textbf{0.9389} & \textbf{0.76}     & \textbf{86.78}  &                                  & \textbf{5.80}      & \textbf{0.9506} & \textbf{0.61}     & \textbf{82.16}  &                                  & \textbf{5.50}      & \textbf{0.9507} & \textbf{0.47}     & \textbf{81.42}  \\ \hline
\multirow{7}{*}{IMDB}             & \textit{Textbugger}              & 81.57                           & 4.18               & 0.9764          & 1.77              & 86.45           & 95.69                            & 3.91               & 0.9788          & 1.63              & 83.85           & 96.08                            & 4.00               & 0.9781          & 1.72              & 85.71           \\
                                  & \textit{Textfooler}              & \underline{77.25}                 & 2.59               & 0.9846          & 0.34              & 78.29           & 95.29                            & 2.46               & 0.9860          & 0.33              & 76.38           & 95.29                            & 2.49               & 0.9858          & 0.33              & 76.73           \\ \cline{3-17} 
                                  & HLA                              & \multirow{5}{*}{\textbf{99.61}} & 16.52              & 0.8790          & 1.75              & 160.33          & \multirow{5}{*}{\textbf{100.00}} & 12.86              & 0.9117          & 1.32              & 150.41          & \multirow{5}{*}{\textbf{100.00}} & 10.75              & 0.9305          & 1.33              & 124.81          \\
                                  & Texthoaxer                       &                                 & 7.96               & 0.9500          & 0.93              & 115.90          &                                  & 4.74               & 0.9721          & 0.56              & 92.87           &                                  & 4.27               & 0.9759          & 0.50              & 88.20           \\
                                  & LeapAttack                       &                                 & 8.96               & 0.9392          & 1.00              & 126.87          &                                  & 4.39               & 0.9743          & 0.54              & 94.79           &                                  & 2.91               & 0.9841          & 0.38              & 79.57           \\
                                  & HQA-Attack                       &                                 & 8.43               & 0.9442          & 0.99              & 124.22          &                                  & 3.96               & 0.9762          & 0.47              & 92.19           &                                  & 2.38               & 0.9883          & 0.34              & 76.60           \\
                                  & \textbf{LBA}                     &                                 & \textbf{2.32}      & \textbf{0.9846} & \textbf{0.29}     & \textbf{77.23}  &                                  & \textbf{2.15}      & \textbf{0.9866} & \textbf{0.24}     & \textbf{75.90}  &                                  & \textbf{1.93}      & \textbf{0.9886} & \textbf{0.19}     & \textbf{74.40}  \\ \hline 
\end{tabular}
}
\end{table*}

\begin{table*}[t]
\caption{Comparison of adversarial texts generated by LBA and other attacks.}
\label{tab:casestudy}
\resizebox{\textwidth}{!}{
\begin{tabular}{c|l|c|c}
\hline
\textbf{Method}                 & \multicolumn{1}{c|}{\textbf{Adversarial Example}}                                                                                                                                                                                                                                                                                                                                                                                                                                                                                                                                                      & \textbf{Target Model}                                                     & \textbf{\begin{tabular}[c]{@{}c@{}}Change of\\ prediction\end{tabular}}                                        \\ \hline
LBA                             & \begin{tabular}[c]{@{}l@{}}\\ Somehow manages to escape the shackles of its own clich s to be the \sout{best} \textcolor{blue}{preferable} espionage picture to come out in weeks .\\ \\ \end{tabular}                                                                                                                                                                                                                                                                                                                                                                                                & \multicolumn{1}{c|}{\multirow{4}{*}{BERT-MR}}                             & \multicolumn{1}{l}{\multirow{4}{*}{\begin{tabular}[c]{@{}l@{}}Positive $\rightarrow$\\ Negative\end{tabular}}} \\ \cline{2-2}
Texthoaxer                      & \begin{tabular}[c]{@{}l@{}}\\ \textcolor{blue}{sorta} manages to \textcolor{blue}{eloped} the \textcolor{blue}{cuffed} of its \textcolor{blue}{singly} clich s to be the \textcolor{blue}{bigger} espionage \textcolor{blue}{depiction} to \textcolor{blue}{became} out in weeks .\\ \\ \end{tabular}                                                                                                                                                                                                                                                                                                 & \multicolumn{1}{l|}{}                                                     & \multicolumn{1}{l}{}                                                                                           \\ \cline{2-2}
LeapAttack                      & \begin{tabular}[c]{@{}l@{}}\\ Somehow \textcolor{blue}{stewardship} to \textcolor{blue}{leakage} the shackles of its own clich s to be the best \textcolor{blue}{spied headshots} to come out in weeks .\\ \\ \end{tabular}                                                                                                                                                                                                                                                                                                                                                                           & \multicolumn{1}{l|}{}                                                     & \multicolumn{1}{l}{}                                                                                           \\ \cline{2-2}
HQA-Attack                      & \begin{tabular}[c]{@{}l@{}}\\ Somehow manages to escape the \textcolor{blue}{strung} of its own clich s to be the \textcolor{blue}{nicer} espionage picture to come out in \textcolor{blue}{date} .\\ \\ \end{tabular}                                                                                                                                                                                                                                                                                                                                                                                & \multicolumn{1}{l|}{}                                                     & \multicolumn{1}{l}{}                                                                                           \\ \hline
LBA                             & \begin{tabular}[c]{@{}l@{}}This restaurant serves really good mexican food at reasonable prices.\\ The location and decor are great and make it look like a really expensive restaurant without the high prices .\\ The service is almost always top \sout{notch} \textcolor{blue}{buttocks} and \sout{prompt} \textcolor{blue}{uneducated}. my pick from the extensive menu would be the enchiladas .\end{tabular}                                                                                                                                                                                    & \multirow{4}{*}{\begin{tabular}[c]{@{}c@{}}LLaMA-7b\\ -chat\end{tabular}} & \multirow{4}{*}{\begin{tabular}[c]{@{}c@{}}Positive $\rightarrow$\\ Negative\end{tabular}}                     \\ \cline{2-2}
Texthoaxer & \begin{tabular}[c]{@{}l@{}}This restaurant serves really good mexican food at reasonable prices .\\ The location and decor are great and make it look like a really \textcolor{blue}{inestimable} restaurant without the high prices .\\ The service is almost \textcolor{blue}{continue} top \textcolor{blue}{gouge} and prompt. my pick from the extensive menu \textcolor{blue}{couldnt} be the enchiladas .\end{tabular}                                                                                                                                                                           &                                                                           &                                                                                                                \\ \cline{2-2}
LeapAttack                      & \begin{tabular}[c]{@{}l@{}}This \textcolor{blue}{dinners attending truthfully adequate wetbacks} food at \textcolor{blue}{wiser award} .\\ The location and decor \textcolor{blue}{becoming marvelous} and \textcolor{blue}{provide} it \textcolor{blue}{consider} like a \textcolor{blue}{very} expensive \textcolor{blue}{cuisine} without the \textcolor{blue}{alto} prices .\\ The service is \textcolor{blue}{approximately siempre paramount} \textcolor{blue}{crack} and prompt . my pick from the \textcolor{blue}{comprehensive journeys} would be the \textcolor{blue}{refried}\end{tabular} &                                                                           &                                                                                                                \\ \cline{2-2}
HQA-Attack                      & \begin{tabular}[c]{@{}l@{}}This \textcolor{blue}{luncheons} serves \textcolor{blue}{altogether commodities pueblo} food at \textcolor{blue}{appropriate} prices .\\ The location and decor are great and make it look like a \textcolor{blue}{surely} expensive \textcolor{blue}{banquets} without the \textcolor{blue}{grands} prices .\\ The service is almost always \textcolor{blue}{topo furrows} and prompt. my pick from the \textcolor{blue}{yawning trajectories got} be the enchiladas .\end{tabular}                                                                                        &                                                                           &                                                                                                                \\ \hline
\end{tabular}
}
\end{table*}
\end{document}